\documentclass[10pt,twocolumn,letterpaper]{article}

\usepackage{amsmath,amsfonts,bm}

\makeatletter
\DeclareRobustCommand\onedot{\futurelet\@let@token\@onedot}
\def\@onedot{\ifx\@let@token.\else.\null\fi\xspace}
\def\cf{\emph{c.f}\onedot}

\def\eg{\emph{e.g}\onedot}

\def\etal{\emph{et al}\onedot}

\def\ie{\emph{i.e}\onedot}

\def\wrt{w.r.t\onedot}
\makeatother

\newcommand{\Eq}{Eq.\@\xspace}

\newcommand*{\myparagraph}[1]{\vspace{0.5em}\noindent\textbf{#1}}

\def\eqref#1{equation~\ref{#1}}

\def\1{\bm{1}}

\DeclareMathAlphabet{\mathsfit}{\encodingdefault}{\sfdefault}{m}{sl}
\SetMathAlphabet{\mathsfit}{bold}{\encodingdefault}{\sfdefault}{bx}{n}

\def\dac{\text{DAC}}
\def\dacv{\text{DAC-V}}

\DeclareMathOperator{\Enc}{\mv{E}}
\DeclareMathOperator{\EncImage}{\Enc\limits_{\text{image}}}
\DeclareMathOperator{\EncText}{\Enc\limits_{\text{text}}}
\DeclareMathOperator{\similarity}{sim}

\newcommand{\data}{\mathcal{D}}
\newcommand{\dataTrain}{\data_{\text{train}}}

\newcommand{\Weights}{\mv{W}}
\newcommand{\WeightsImage}{\Weights_{\text{image}}}
\newcommand{\WeightsText}{\Weights_{\text{text}}}

\newcommand{\Labels}{\mv{L}}
\newcommand{\LabelsOneHot}{\Labels_{\text{one\_hot}}}

\newcommand{\zimage}{\mv{z}_{\text{image}}}
\newcommand{\ztext}{\mv{z}_{\text{text}}}

\newcommand{\vimage}{\mv{v}_{\text{image}}}
\newcommand{\vtext}{\mv{v}_{\text{text}}}

\newcommand{\WeightsImageDAC}{\WeightsImage^{\dac}}
\newcommand{\WeightsTextDAC}{\WeightsText^{\dac}}

\newcommand*{\mv}[1]{\mathbf{#1}}

\newcommand*{\norm}[1]{\lVert#1\rVert}
\DeclareMathOperator{\reals}{\mathbb{R}}

\newcommand{\linearlayer}{\mv{H}}

\usepackage[algorithms,pagenumbers]{wacv}

\usepackage{graphicx}
\usepackage{amsmath}
\usepackage{amssymb}
\usepackage{booktabs}
\usepackage{mathtools}
\usepackage{multirow}

\usepackage{nicefrac} %
\usepackage{balance} %
\usepackage[T1]{fontenc} %

\usepackage[pagebackref,breaklinks,colorlinks]{hyperref}
\usepackage[accsupp]{axessibility}  %

\usepackage[capitalize]{cleveref}
\crefname{section}{Sec.}{Secs.}
\Crefname{section}{Section}{Sections}
\Crefname{table}{Table}{Tables}
\crefname{table}{Tab.}{Tabs.}

\def\wacvPaperID{} %
\def\confName{WACV}
\def\confYear{2024}

\usepackage{setspace}
\usepackage{fancyhdr}

\begin{document}

\title{Domain Aligned CLIP for Few-shot Classification}

\author{
    Muhammad Waleed Gondal, Jochen Gast, Inigo Alonso Ruiz, Richard Droste,
    \\
    Tommaso Macri, Suren Kumar, Luitpold Staudigl
    \\
    Amazon\\
    {\tt\small \{wgondal, jogast, inruiz, rdroste, tmacri, ssurkum, luitpold\}@amazon.com}
}

\maketitle
\thispagestyle{fancy} %

\begin{abstract}
Large vision-language representation learning models like CLIP have demonstrated impressive performance for zero-shot transfer to downstream tasks while largely benefiting from inter-modal (image-text) alignment via contrastive objectives.
This downstream performance can further be enhanced by full-scale fine-tuning which is often compute intensive, requires large labelled data, and can reduce out-of-distribution (OOD) robustness.
Furthermore, sole reliance on inter-modal alignment might overlook the rich information embedded within each individual modality.
In this work, we introduce a sample-efficient domain adaptation strategy for CLIP, termed Domain Aligned CLIP (DAC), which improves both intra-modal (image-image) and inter-modal alignment on target distributions without fine-tuning the main model.
For intra-modal alignment, we introduce a lightweight adapter that is specifically trained with an intra-modal contrastive objective.
To improve inter-modal alignment, we introduce a simple framework to modulate the precomputed class text embeddings.
The proposed few-shot fine-tuning framework is computationally efficient, robust to distribution shifts, and does not alter CLIP's parameters.
We study the effectiveness of DAC by benchmarking on 11 widely used image classification tasks with consistent improvements in 16-shot classification upon strong baselines by about 2.3\% and demonstrate competitive performance on 4 OOD robustness benchmarks.
\end{abstract}

\section{Introduction}
\label{intro}
The growing popularity of large-scale representation learning models for multi-modal data like CLIP \cite{radford2021learning}, ALIGN \cite{jia2021scaling} and Florence \cite{yuan2021florence} has highlighted the need for efficiently adapting these models to various downstream tasks across multiple domains and applications.
In CLIP, for instance, successful zero-shot transfer relies on adapting to both image and text modalities in new domains and aligning their representations (inter-modal alignment) in a shared representation space.
However, the generalization capability of CLIP is constrained by its pre-training distribution \cite{fang2022data}.
To enhance their performance, these models are often transferred to target distributions either through fine-tuning or employing various few-shot strategies. While fine-tuning can be resource intensive and prone to overfitting \cite{pham2021combined, wortsman2022robust}, few-shot adaptation offers a training and sample efficient alternative.
\begin{figure*}[t]
\begin{minipage}[b]{0.57\linewidth}
\centering
\includegraphics[width=\linewidth]{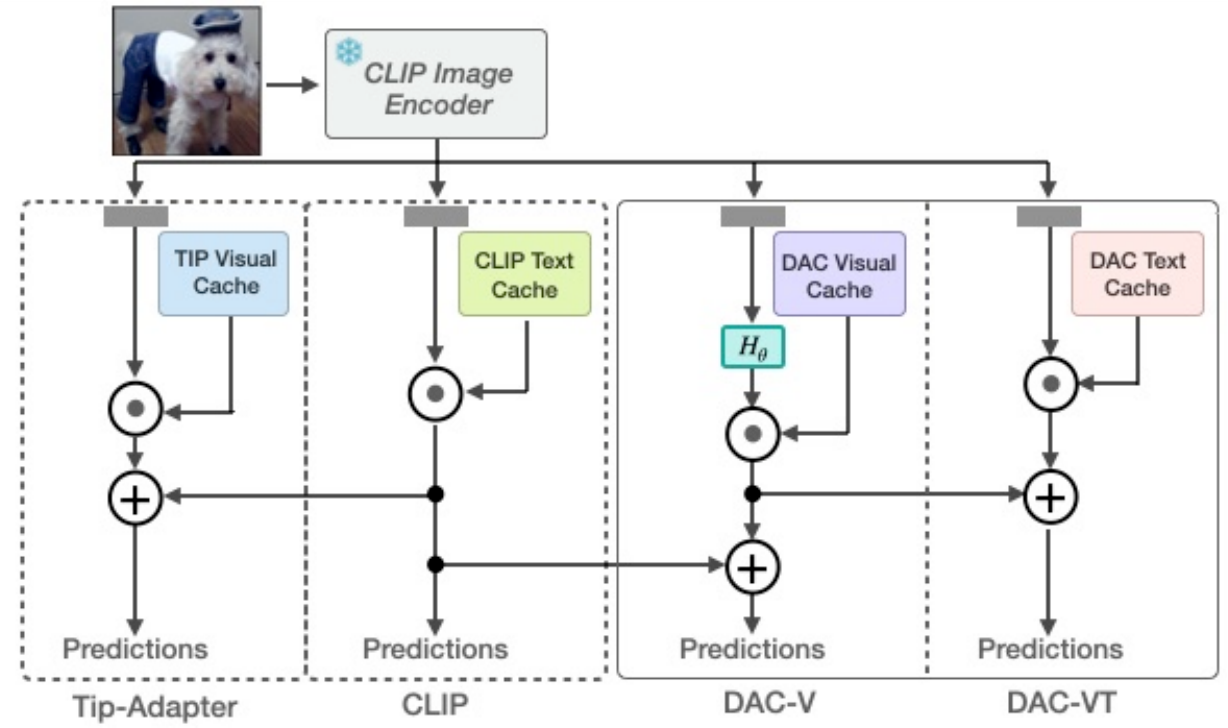}
\captionsetup{labelformat=empty}
\caption*{(a) Overview of test-time inference for different methods.}
\end{minipage}
\begin{minipage}[b]{0.42\linewidth}
\centering
\includegraphics[width=1.0\linewidth]{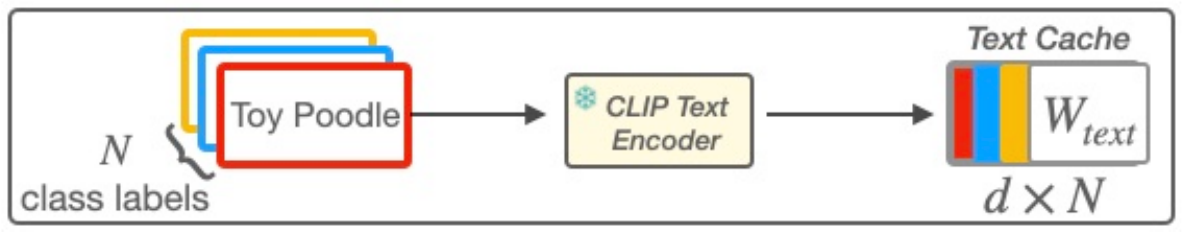}
\captionsetup{labelformat=empty}
\caption*{(b) Text Cache Construction.}
\hspace{0.05\linewidth}
\includegraphics[width=1.0\linewidth]{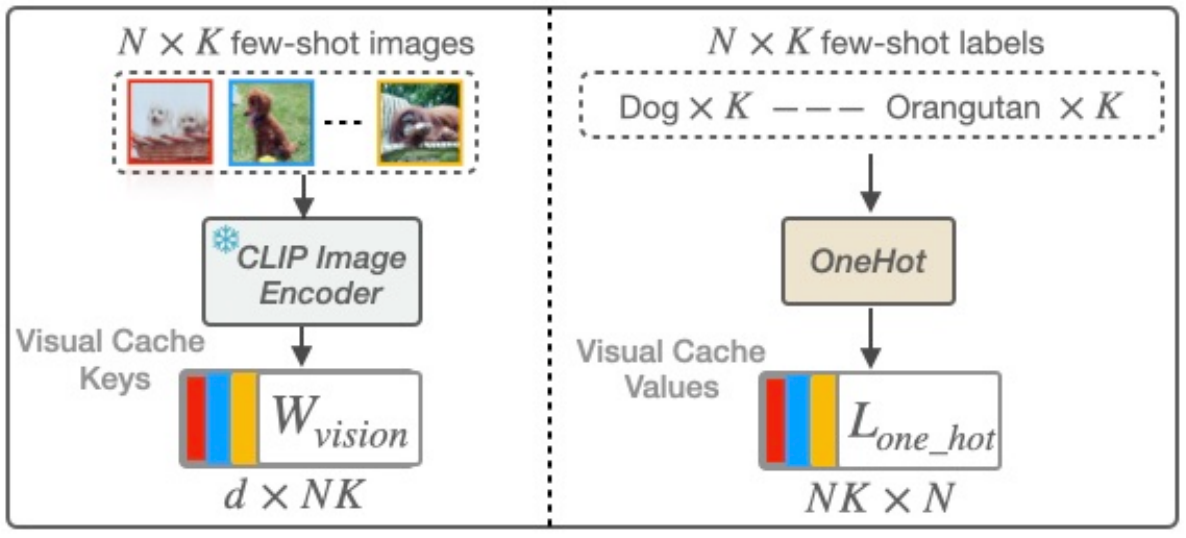}
\captionsetup{labelformat=empty}
\caption*{(c) Visual Cache Construction.}
\end{minipage}
\caption{%
{\bf Overview of CLIP and its few-shot adaptations}.
(a) To make zero-shot predictions, CLIP uses precomputed text embeddings of class labels (termed as text cache) to determine inter-modal similarities between images and text embeddings. 
Tip-Adapter extends CLIP for few-shot prediction by compounding its inter-modal logits with intra-modal logits. 
The intra-modal logits stem from precomputed image embeddings of a few labeled images (termed as visual cache). 
DAC-V adapts image embeddings of the visual cache to the target distributions. 
In addition to the adaptation of image embeddings, DAC-VT adapts textual embeddings on the target distributions. 
The construction of text and visual caches from a few seen examples are shown in (b) and (c), respectively.}
\label{fig:methods_overview}
\end{figure*}

In the few-shot setup, in addition to the class labels, we are also provided with a few labeled images from target distributions.
These labeled images serve as data-specific priors which can be used to update CLIP's existing inter-modal predictions.
In this work, we present an approach called DAC (Domain Aligned CLIP) that effectively leverages this prior knowledge to adapt CLIP for new downstream tasks.
We posit that a combination of improved intra-modal (image-image) and inter-modal (image-text) alignment in the target domain results in better few-shot transfer to downstream classification tasks.
To achieve this, we split the overall classification task into an ensemble of intra- and inter-modal classifications.
While the inter-modal classification leverages image-text similarity (as in CLIP), the intra-modal classification is performed by means of a visual cache that is composed of precomputed image embeddings of the few (seen) labelled images, \cf \cref{fig:methods_overview}. 

\begin{figure}%
    \centering
    \begin{subfigure}{0.9\linewidth}
        \begin{subfigure}[t]{0.98\textwidth}
            \includegraphics[width=\textwidth]{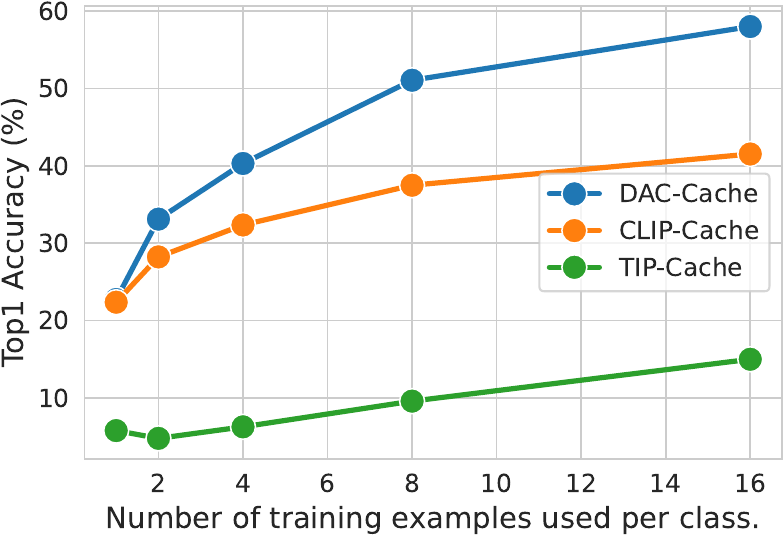}
        \end{subfigure}

    \end{subfigure}
\caption{{\bf Comparative analysis of different intra-modal classifiers using 16-shots visual-cache}. 
Here, TIP-Adapter's cache looses the intra-modal classification advantage of CLIP. 
In contrast, the intra-modal classifier based on DAC visual features performs consistently better.
} 
\label{fig:intra_vs_inter}
\end{figure}

A similar ensembling framework was recently leveraged in Tip-Adapter \cite{zhang2021tip} for few-shot CLIP adaptation. However, no explicit regularization is done to improve the intra-modal alignment. While keeping the inter-modal classification fixed, Tip-Adapter(-F) treats the visual cache as learnable parameters and optimizes them to learn the residual information required to improve the upstream classification performance.
We show that such an optimization causes the visual cache to lose its diverse, rich visual information and deteriorates its discriminative capability as depicted by its intra-modal classification performance \cf \cref{fig:intra_vs_inter}. Hence, despite relying on an ensemble to exploit
feature diversity, Tip-Adapter-F reduces this diversity and limits feature reuse which is crucial for transfer learning and robustness \cite{neyshabur2020being}. Furthermore, Tip-Adapter-F does not adapt textual features in the target domain which can be crucial as recent work highlights the limitations of CLIP for inter-modal alignment \cite{liang2022mind} and how performance degrades as downstream vocabulary expands \cite{ren2022rethinking}.

In this work, unlike Tip-Adapter(-F), we introduce a two stage adaptation strategy that focuses on individually improving both intra- and inter-modal classifiers. See \cref{fig:methods_overview} for an overview of methods. Our hypothesis is grounded in the widely acknowledged phenomenon that effective ensembles consist of models that are both accurate and make uncorrelated errors \cite{opitz1999popular, gontijo2021no, wen2020batchensemble}. In the first stage of the proposed setup, a linear adapter layer is exclusively trained by a (self-)supervised contrastive objective to contrast images sampled from different classes. The goal is to improve the affinity of the latent representations of images coming from the same class while keeping representations of images belonging to different classes distant. This results in an improved intra-modal classifier which forms the basis of DAC-V. In the second stage, we introduce a framework that directly optimizes CLIP's text embeddings and improves inter-modal classifier performance while ensembling it with the frozen intra-modal classifier from the first stage. We call this overall framework DAC-VT where both visual and textual representations are adapted to the target distribution. 

Our \textbf{primary contributions} are as follows:
\begin{itemize}
    \item We present DAC, a novel framework for adapting CLIP for few-shot classification tasks that learns by explicitly aligning intra-modal and inter-modal representations on target distributions. To the best of our knowledge, this is the first work to leverage intra-modal regularization for few-shot adaptation of large vision-language models.
    \item We conduct comprehensive quantitative analysis on 11 widely used image classification benchmarks and show that our method outperforms competitive baselines, while maintaining reasonable robustness to distributions shifts (measured on 4 benchmarks).
\end{itemize}
\section{Related Work}
\label{sec:related_work}

\noindent
Learning rich representations of data that generalize well to multiple tasks is challenging but desirable \cite{bengio2013representation}. Such representations not only enable sample-efficient transfer to downstream tasks but also simplify the hyper-parameter optimization \cite{kolesnikov2020big}.
To this end, contrastive learning \cite{gutmann2010noise, oord2018representation} based self-supervised methods have shown to be promising for learning transferable representations of images \cite{tian2020contrastive, he2020momentum, chen2020simple} and text \cite{oord2018representation}. More recently, these objectives have been extended to align data from different modalities in a joint representation space \cite{radford2021learning,jia2021scaling,yuan2021florence,duan2022multi,yang2022vision}, achieving impressive zero-shot transfer learning performance on a number of downstream tasks.
The performance of these models can further be improved by either fine-tuning them on the labeled target data \cite{wortsman2022robust,radford2021learning} or by adapting their feature space on the target distribution while maintaining feature re-useability \cite{neyshabur2020being}. The second strategy is practically more appealing, as it is sample efficient and only requires simple hyperparameter tuning \cite{houlsby2019parameter}. For CLIP \cite{radford2021learning}, such sample-efficient adaptation methods can be broadly classified into two categories; (1) methods that learn to optimize text prompts \cite{zhou2022learning, zhang2021vt, zhou2022conditional, ma2022understanding,yao2021cpt}, and, (2) methods that introduce lightweight adapter layers to align image and text embeddings on target distributions \cite{gao2021clip, peng2022sgva, zhang2021tip, guo2022calip}. The latter setting offers more flexibility in adapting both visual and textual domains, where CLIP-Adapter \cite{gao2021clip} tunes adapter layers appended to CLIP's frozen image and text encoders.

\noindent
\textbf{Differences to CLIP-Adapter:} DAC differs from CLIP-Adapter in three ways: (1) DAC explicitly optimizes the intra-modal alignment of visual features, whereas optimization in CLIP-Adapter is solely geared towards inter-modal alignment. (2) DAC only uses a single linear-layer adapter for intra-modal alignment, whereas CLIP-Adapter uses two-layer MLPs on both visual and textual features for inter-modal alignment. (3) Inter-modal alignment in DAC is done by directly modulating the text cache which is much more efficient in comparison to using a separate adapter for text embeddings (as in CLIP-Adapter).

\noindent
\textbf{Differences to Tip-Adapter:}
Tip-Adapter \cite{zhang2021tip} employs a visual caching structure to split the overall classification into an ensemble of intra-modal and inter-modal classifiers. However, note that the ensembling in Tip-Adapter yields inefficient sub-classifiers, limiting the full utilization of few-shot knowledge available. We propose two ways to improve it. First, we introduce an intra-modal contrastive learning framework to improve the visual alignment of features in the target domain. We extend the function contrastive objective used in \cite{gondal2021function} to approximate the proxy visual function-space of downstream classes. Second, we fine-tune CLIP's precomputed textual embeddings to mitigate its limitations with unseen vocabulary \cite{ren2022rethinking} for class labels in the target domain. The framework is simpler than prompt tuning and does not require any additional parameters. Moreover, unlike Tip-Adapter, our method eliminates the need for an extra sharpness parameter for tuning image similarity scores.

While efforts have been made to adapt CLIP features without fine-tuning at test-time at the cost of reduced in-distribution performance \cite{shu2022test,udandarao2022sus}, we specifically focus on few-shot fine-tuning for CLIP adaptation, noting that these methods complement our work. Additionally, recent work leverages pre-trained language models to generate additional category information \cite{xiao2022exploiting, zhang2023prompt} and visual generative models to synthesize images for expanding few-shot training data \cite{zhang2023prompt}. Unlike these methods, DAC only uses the few-shot data provided with the task, making a comparison with these methods unfair.
\section{Background}
\label{sec:background}
We start by explaining CLIP and how its zero-shot prediction can be formulated by means of a \textit{text-cache}.
We then expand on how this formulation is extended in Tip-Adapter to support few-shot classification tasks via a \textit{visual-cache}.

\myparagraph{Zero-shot Classification with a Text-Cache.}
\label{sec:clip-zero-shot}
CLIP is a vision-language representation learning model that aligns vision and text modalities in a joint embedding space by learning from image-text pairs $(\mv{x}, \mv{t})$ where $\mv{x}$ are vectorized images and $\mv{t}$ correspond to tokenized text inputs.
At inference-time, CLIP encodes these input modalities into \mbox{$d$-dimensional} embeddings with separate encoders for image and text, \ie 
$\vimage = \EncImage(\mv{x})$ and $\vtext = \EncText(\mv{t})$.
For brevity, we will refer to the L2-normalized embeddings as 
$\zimage = \nicefrac{\vimage}{\norm{\vimage}}$ and $\ztext = \nicefrac{\vtext}{\norm{\vtext}}$.
Alignment between image and text embeddings is then computed via cosine similarity, 
\ie $\similarity (\vimage,\vtext) := \vimage^T \vtext^{} / (\norm{\vimage} \norm{\vtext}) = \zimage^T \ztext^{}.$
\vspace{0.1em}
CLIP leverages this image-text alignment for zero-shot classification with novel inputs.
Assume that a given task consists of $N$ classification labels $\{y^{(i)}\}_i^N$.
We first construct a precomputed weight matrix (or \textit{text-cache}) by concatenating (normalized) text embeddings of all classification labels.
\begin{align}\label{eq:textcache}
\WeightsText &= \begin{pmatrix} \ztext^{(1)} & \ztext^{(2)} & \ldots & \ztext^{(N)} \end{pmatrix} \in \reals^{d \times N},
\end{align}
which encapsulates pre-computed, textual knowledge associated with the task.
Subsequently, the text-cache $\WeightsText$ can be used to classify a new, unseen input image into $N$ classes by computing \textit{inter-modal} logits, \ie
\begin{align}\label{eq:clip-logits}
\mv{\text{logits}}_{\text{CLIP}} = \WeightsText^T \, \zimage^{} \in \reals^{N \times 1}.
\end{align}
Note that $\WeightsText$ only needs to be computed once per task.

\myparagraph{Few-shot Classification with a Visual-Cache.}
\label{sec:tip-adapter}
Tip-Adapter \cite{zhang2021tip} extends CLIP for few-shot classification. 
For each new task, it requires a few labeled training examples from a target distribution $\dataTrain = \{(x^{(i)}, y^{(i)})\}_i^{N \times K}$ where $N$ is the number of classes and $K$ is the number of examples (or shots) per class. 
Tip-Adapter encodes this few-shot knowledge into a precomputed \textit{visual-cache} with separate cache keys and values.
Akin to \cref{eq:textcache}, cache keys are computed as
\def\matrixspacinga{\,\,\,\,}
\def\matrixspacingb{\,}
\begin{align}\label{eq:visualcache}
\WeightsImage &= \Big( \matrixspacingb \zimage^{(1,1)} \matrixspacinga \zimage^{(1,2)} \matrixspacinga \ldots \matrixspacinga \zimage^{(K,N)} \matrixspacingb \Big)
\in \reals^{d \times NK},
\end{align}
where $\WeightsImage$ is the concatenation of (sub-)weight matrices 
$\begin{psmallmatrix} \zimage^{(1, \cdot)} & \zimage^{(2, \cdot)} & \ldots & \zimage^{(K, \cdot)} \end{psmallmatrix}$ 
per classification label horizontally.
Corresponding cache values are then constructed as one-hot encodings $\LabelsOneHot \in \mathbb{R}^{N K \times N}$ of ground truth labels $\{y_i\}_i^{N \times K}$ by vertically concatenating one-hot encodings per shot, followed by horizontal concatenation per classification label.
Note that such a key-value configuration effectively enables the visual-cache to retain all the available few-shot knowledge in $\dataTrain$. 
To update CLIP logits (\Eq\ref{eq:clip-logits}) with the few-shot knowledge encoded in the visual-cache (\Eq\ref{eq:visualcache}), Tip-adapter introduces an affinity vector
\begin{align}\label{eq:tip_affinity}
\mv{w}_{\text{affinity}} &= \exp \big( \beta \, (\WeightsImage^T \, \zimage^{}  - 1 )\big) \in \reals^{N K \times 1},
\end{align}
where $\exp$ denotes a pointwise exponential function and $\beta$ modulates the sharpness of affinities.
The affinity vector $\mv{w}_{\text{affinity}}$ retains the similarity (or compatibility) between a given image $\zimage$ and the images stored in the visual-cache $\WeightsImage$.
Tip-Adapter finally computes aggregated logits as
\begin{align}\label{eq:tip-final-logits}
\mv{\text{logits}}_{\text{TIP}} = \mv{\text{logits}}_{\text{CLIP}} + \alpha \,\,  \LabelsOneHot^T \, \mv{w}_{\text{affinity}}^{},
\end{align}
where the second term denotes \textit{intra-modal} logits. 
Here, the few-shot knowledge in the visual-cache is used to update CLIP's inter-modal predictions, \cf \cref{fig:methods_overview}.
Note that $\alpha$ trades off contributions of the visual and text-cache towards the final prediction. 
Furthermore, \cite{zhang2021tip} proposes Tip-Adapter-F which improves upon Tip-Adapter by optimizing the visual-cache $\WeightsImage$ \wrt $\dataTrain$ to learn the residual information required to increase the upstream classification performance in the target domain.

\section{Domain Aligned CLIP}
\label{sec:method}

\begin{figure}
    \centering
    \begin{subfigure}{1\linewidth}
        \includegraphics[width=\textwidth]{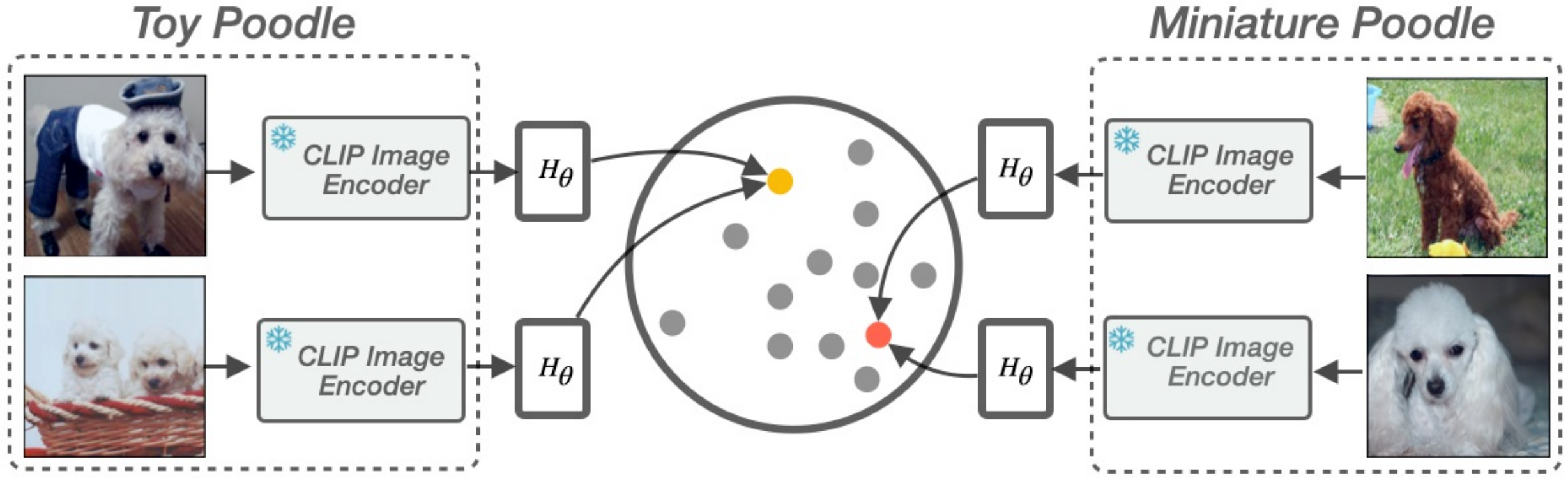}
    \end{subfigure}
    \caption{{\bf Construction of supervised contrastive objective used to fine-tune the visual adapter layer in DAC.}}
    \label{fig:contrastive_objective}
    \vspace{-1.em}
\end{figure}
We now introduce Domain Aligned CLIP (DAC), a method that improves the few-shot domain adaptation of CLIP in two stages. 
In the first stage, we tune a visual adapter layer to align CLIP's visual representation in the target distribution, resulting in an improved intra-modal classifier.
This intra-modal classifier later becomes the basis of DAC-V. 
In the second stage, we fine-tune CLIP's textual representation for improved inter-modal alignment in the target distribution. 
This inter-modal classifier, together with the intra-modal classifier from the first stage results in DAC-VT. 
See \cref{fig:methods_overview} for an overview of both methods. %

\subsection{Adapting the Visual Domain}
\label{sec:adapting-visual-domain}
Tip-Adapter shows how an inter-modal classifier based on a text-cache can be improved when ensembled with a visual-cache based intra-modal classifier.
However, the performance of the intra-modal classifier is inferior to the inter-modal classifier.
This is due to CLIP being explicitly trained for inter-modal alignment between images and text where it is not explicitly encouraged to align embeddings of images sharing the same underlying concept or class.
Moreover, recent work \cite{duan2022multi, yang2022vision} has shown the benefits of enforcing intra-modal alignment in pre-training of CLIP-like models.

In this work, we do not train models from scratch.
Instead, we rely on a few labeled examples from the target domain to enhance the intra-modal alignment of pre-trained CLIP models without affecting their inter-modal alignment.

\myparagraph{DAC Visual Adapter.} 
To align the visual features of CLIP in the target domain, we introduce a linear layer $\linearlayer_{\bm{\theta}}$ as an adapter that is appended to the frozen CLIP image encoder. 
During visual adaptation we only fine-tune the parameters $\bm{\theta}$.
Unlike the two-layered adapter in \cite{gao2021clip}, we found that a single linear layer is effective and avoids over-fitting. 
To allow the unimpeded passage of features at the beginning of fine-tuning, we initialize $\linearlayer_{\bm{\theta}}$ as the identity, making it stable and efficient \cite{houlsby2019parameter}. %

\begin{figure}%
    \centering
    \begin{subfigure}[t]{1.0\linewidth}
        \includegraphics[width=\textwidth]{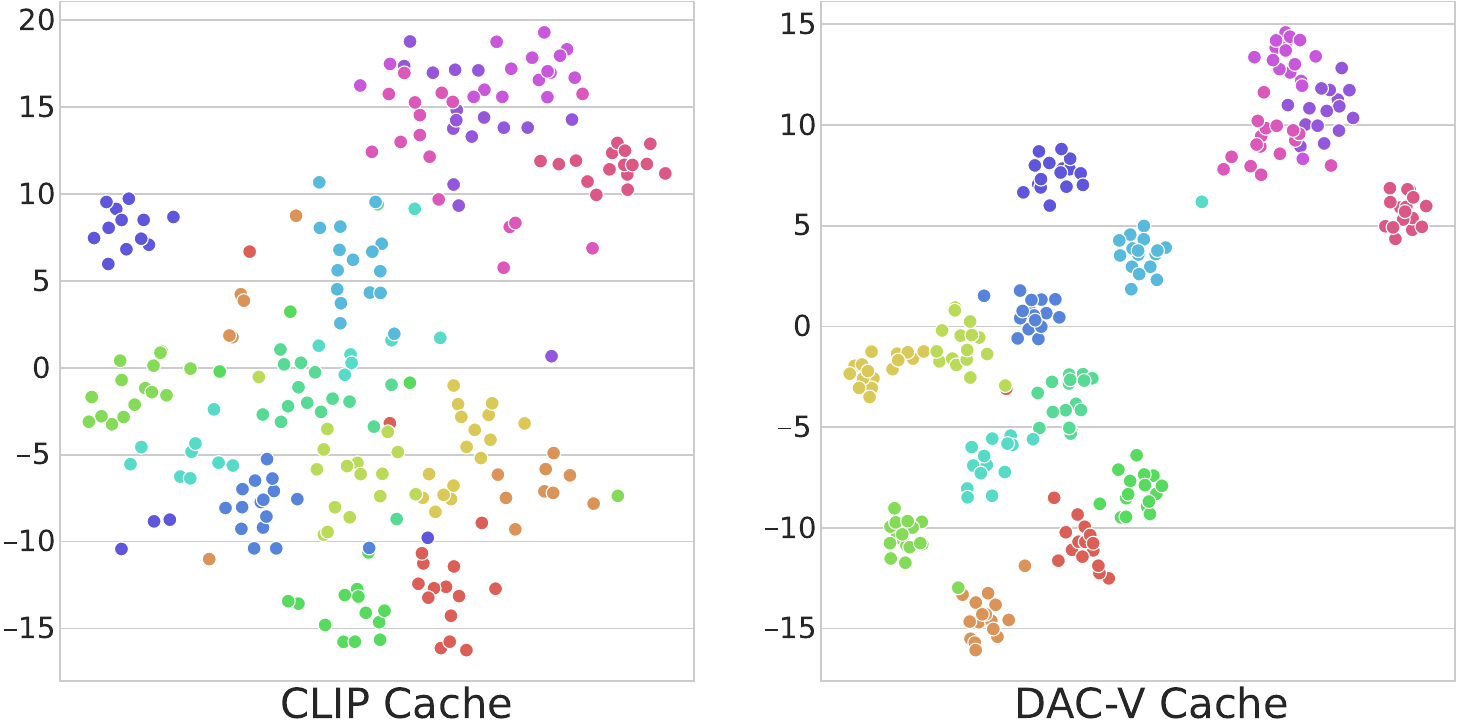}
    \end{subfigure}
    \caption{{\bf 2D tSNE projections of the CLIP's and DAC-V's cached image embeddings}. Our intra-modal constrastive objective enforces structure on the representations.} 
    \label{fig:tsne}
\vspace{-1.em}
\end{figure}

\myparagraph{Visual Adapter Training.} 
Next, we train the adapter layer $\linearlayer_{\bm{\theta}}$ to minimize the distance between image embeddings of the same class in the latent space while pushing them apart for images of different classes. 
Similar to \cref{sec:tip-adapter}, we assume a few-shot setting with a novel target distribution $\mathcal{D}_\text{train}$ given $N$ classes and $K$ shots.
To ensure that all $K$ images of the same class are mapped to similar representations, we formulate a supervised contrastive objective, as illustrated in \cref{fig:contrastive_objective}. 
Here, we consider $K$ images as the context set $\mathcal{C}_s$ of a parent class $s \in \{1,\dots,N\}$ which is further supplemented with $M$ randomly augmented views of the given images,
\ie  $\mathcal{C}_s = \{(\mv{x}^{(i)}_s, \mv{y}^{(i)}_s)\}_{i}^{K \times M}$. 
In \cref{sec:ablations} we provide ablations for choosing an appropriate number of augmented views.
To apply the visual adaptation, we linearly transform the image embeddings obtained from the frozen CLIP image encoder
$\mv{v}^{\bm{\theta}}_{\text{image}} = \linearlayer_{\bm{\theta}} \EncImage(\mv{x})$,
followed by L2-normalization
$\mv{g} = \nicefrac{\vimage^{\bm{\theta}}}{\norm{\vimage^{\bm{\theta}}}}$. %
Note that we drop the dependency on $\bm{\theta}$ for brevity.
We aim to find the optimal transformation by minimizing the contrastive loss
\newcommand{\mylocalv}[2]{\bm{\vartheta}^{(#2)}_{\text{image},#1}}
\newcommand{\mylocalg}[2]{\mv{g}^{(#2)}_{#1}}
\begin{align}\label{eq:visual_contrastive}
\sum\limits_{n=1}^{N}\sum\limits_{1\le i<j\le MK} %
\log \frac{ \exp \left[ (\mylocalg{j}{n})^T \mylocalg{i}{n} \, / \tau \right] }{%
\sum_{q=1}^{N} \exp \left[ (\mylocalg{j}{n})^T \mylocalg{i}{q} \, / \tau \right]},
\end{align}
where $\tau$ is a temperature to scale cosine similarities.
Minimizing \cref{eq:visual_contrastive} aims to maximize the similarity between embeddings pairs coming from the same class (\textit{positive} pairs), while maximizing dissimilarity between embedding pairs of different classes (\textit{negative} pairs).
Note that the summation $1 \le i < j \le MK$ in \cref{eq:visual_contrastive} considers all the positive pairs in $\mathcal{C}_N$, a total of $MK \choose 2$ combinations. 
Our visual adaptation enforces structure onto the visual embeddings which is demonstrated by
a much better cluster separation; \cf \cref{fig:tsne}.

\myparagraph{Constructing DAC-V.}
Leveraging the learned transformation $\linearlayer_{\bm{\theta}}$, we can improve \cref{eq:visualcache} by an adapted visual-cache
\def\matrixspacingaa{\,\,\,}
\def\matrixspacingbb{\,}
\def\gimage{\mv{g}}
\begin{align}\label{eq:weights-image-dac}
\WeightsImageDAC &= \Big( \matrixspacingbb \gimage^{(1,1)} \matrixspacingaa \gimage^{(1,2)} \matrixspacingaa \ldots \matrixspacingaa \gimage^{(K,N)} \matrixspacingbb \Big)
\in \reals^{d \times NK},
\end{align}
where we apply a horizontal concatenation as before.
By inserting the improved visual-cache (\Eq~\ref{eq:weights-image-dac}) into \cref{eq:tip_affinity}, we obtain an optimized affinity vector that is visually adapted to the given task.
However, note that the parameter $\beta$ is subsumed into the learnable linear transformation. 
Hence, we have the optimized affinity vector
\begin{align}\label{eq:dacv-affinity}
\mv{w}^{\dacv}_{\text{affinity}} &= \exp \big( (\WeightsImageDAC)^T \, \mv{g}_{\text{image}}  - 1 \big) \in \reals^{N K \times 1}.
\end{align}
While \cref{eq:tip_affinity} and \cref{eq:dacv-affinity} share similarities, they differ in a crucial aspect.
That is, by introducing a learned linear transformation in \cref{eq:weights-image-dac,eq:dacv-affinity}, the intra-modal representation of DAC-V gets tailored towards the novel task, while \cref{eq:tip_affinity} remains static and does not perform such domain adaptation.  
Similar to \cref{eq:tip-final-logits}, we obtain the final logits as
\begin{align}\label{eq:dacv-final-logits}
\mv{\text{logits}}_{\text{DAC-V}} = \mv{\text{logits}}_{\text{CLIP}} + \alpha \,\,  \LabelsOneHot^T \, \mv{w}^{\dacv}_{\text{affinity}}.
\end{align}
In contrast to \cref{eq:tip-final-logits}, the second term in \cref{eq:dacv-final-logits} is composed of image features that are visually aligned in the target distribution.
Our experiments demonstrate that DAC-V, on average, outperforms the fine-tuned Tip-Adapter-F on 11 image benchmarks by 0.83\%, \cf \cref{sec:experiments}.

\subsection{Adapting the Textual Domain}
\label{sec:text_alignment}
Having optimized the intra-modal representations in DAC-V, we now look at enhancing inter-modal alignment between images and text features in the target distribution. 
Previous work \cite{zhou2022learning, gao2021clip} demonstrates the benefits of optimizing text embeddings for few-shot classification through the optimizing of text prompts, while concurrently keeping the attached class names fixed. 
Such a prompt-tuning framework, however, significantly lacks the flexibility of fine-tuning the text embeddings. There are two severe issues:
First, it does not address the adaptation of CLIP's vocabulary to the new class names from a target distribution as CLIP's learned vocabulary is shown to have limitations \cite{ren2022rethinking}. 
Second, there may be multiple visual concepts associated with different class names that can cause confusion among competing text features of CLIP \cite{tsipras2020imagenet, beyer2020we}; see \cref{sec:aligning_textual_downstream_tasks}.
To circumvent these challenges, we present a significantly simpler framework to align images and text labels in the target distribution.

\myparagraph{DAC Textual Adapter.}
Unlike previous work, we do not introduce a new adapter module for improving inter-modal alignment. Instead, we directly fine-tune the text-cache $\WeightsText$. 
Note that the text embeddings are continuous vectors that encapsulate the concepts specified by class names from a target distribution.
Therefore, modulating them influences the overall class description.

\myparagraph{Constructing DAC-VT.}
To ensure smooth integration of the inter-modal alignment with the previously proposed intra-modal classifier in \mbox{DAC-V}, we optimize $\WeightsText$ in the ensembled setting.
More specifically, we convert $\WeightsText$ into a learnable vector and freeze all the remaining components including the visual cache components and $\linearlayer_{\bm{\theta}}$. 
Thereafter, using the few-shot dataset, we optimize $\WeightsText$ to align text embeddings with the visual embeddings while keeping the weighting parameter fixed,
\ie $\alpha = 1$.
The optimized text weights $\WeightsTextDAC$ result in the DAC-VT classifier, \ie
\begin{align}\label{eq:dacvt-final-logits}
\mv{\text{logits}}_{\text{DAC-VT}} = (\WeightsTextDAC)^T \zimage^{} + \alpha \,\,  \LabelsOneHot^T \, \mv{w}^{\dacv}_{\text{affinity}}.
\end{align}
Intuitively, optimizing the inter-modal alignment in this ensembled setting encourages $\WeightsTextDAC$ to assimilate the prior few-shot knowledge acquired by DAC-V. In \cref{sec:ablations}, we ablate over other ways of constructing DAC-VT (including an end-to-end setting) that results into sub-optimal ensembles of intra-modal and inter-modal classifiers.

\section{Experiments}
\label{sec:experiments}
In this section, we quantitatively evaluate our proposed method on 11 commonly used image classification tasks. We also study its robustness to distribution shifts. In \cref{sec:ablations}, we ablate over DAC components and the design choices.

\begin{figure*}[!h]
    \centering

    \begin{subfigure}{1\textwidth}
        \begin{subfigure}{0.245\textwidth}
            \includegraphics[width=\textwidth]{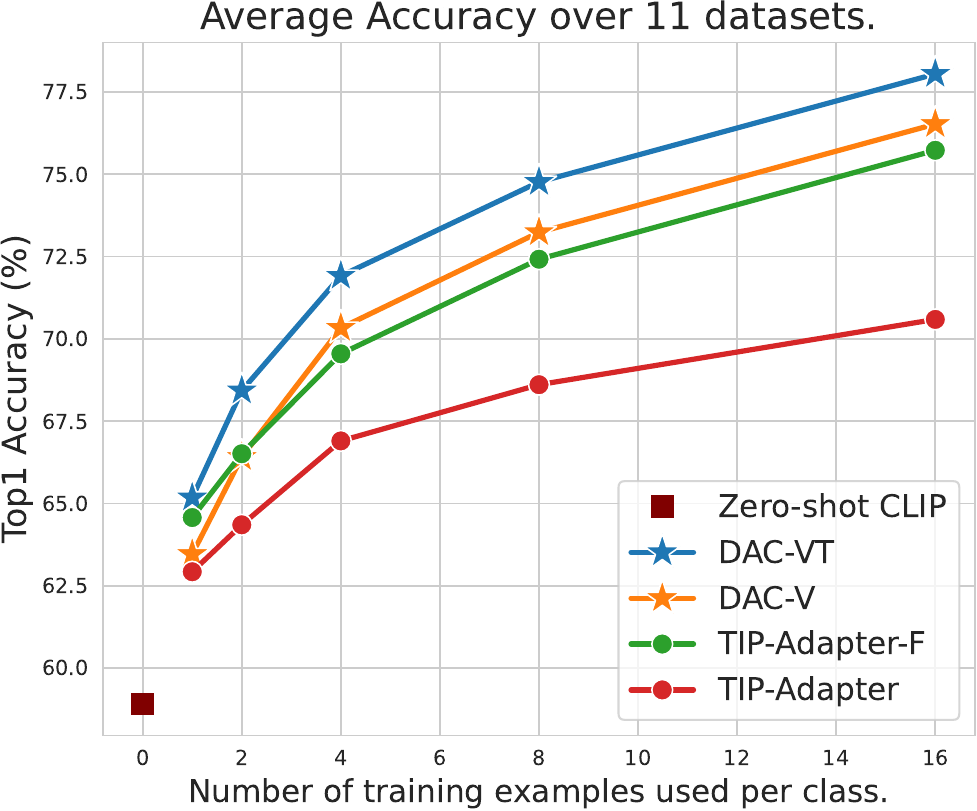}
        \end{subfigure}
        \begin{subfigure}{0.245\textwidth}
            \includegraphics[width=\textwidth]{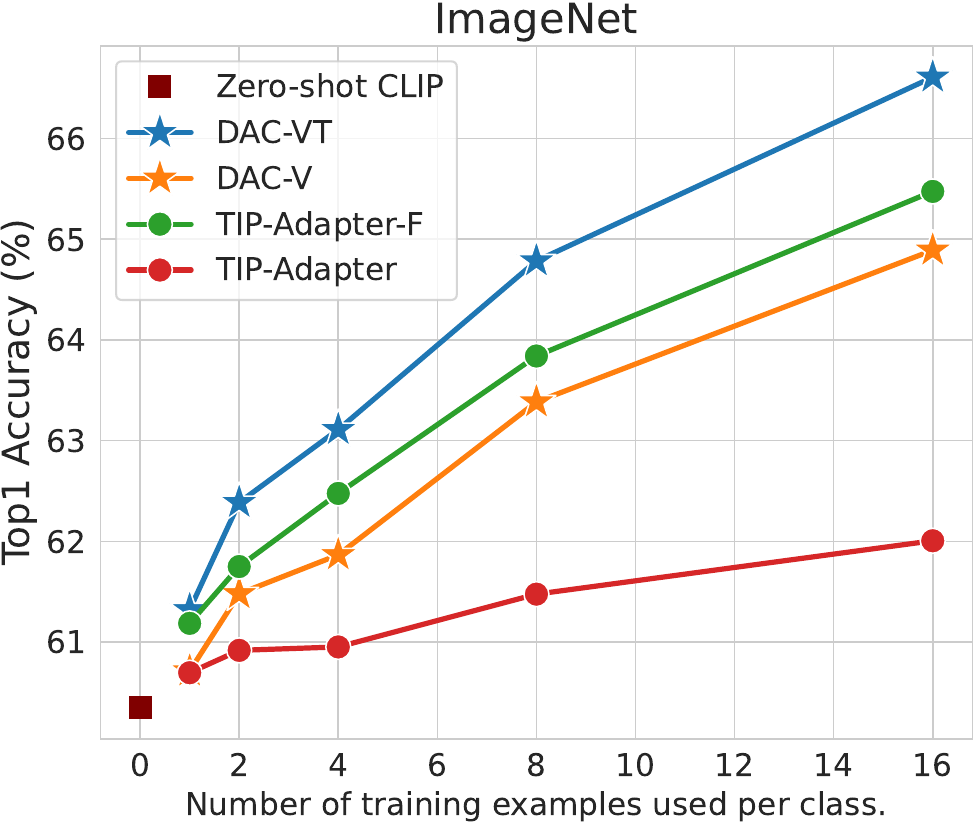}
        \end{subfigure}%
        \begin{subfigure}{0.245\textwidth}
            \includegraphics[width=\textwidth]{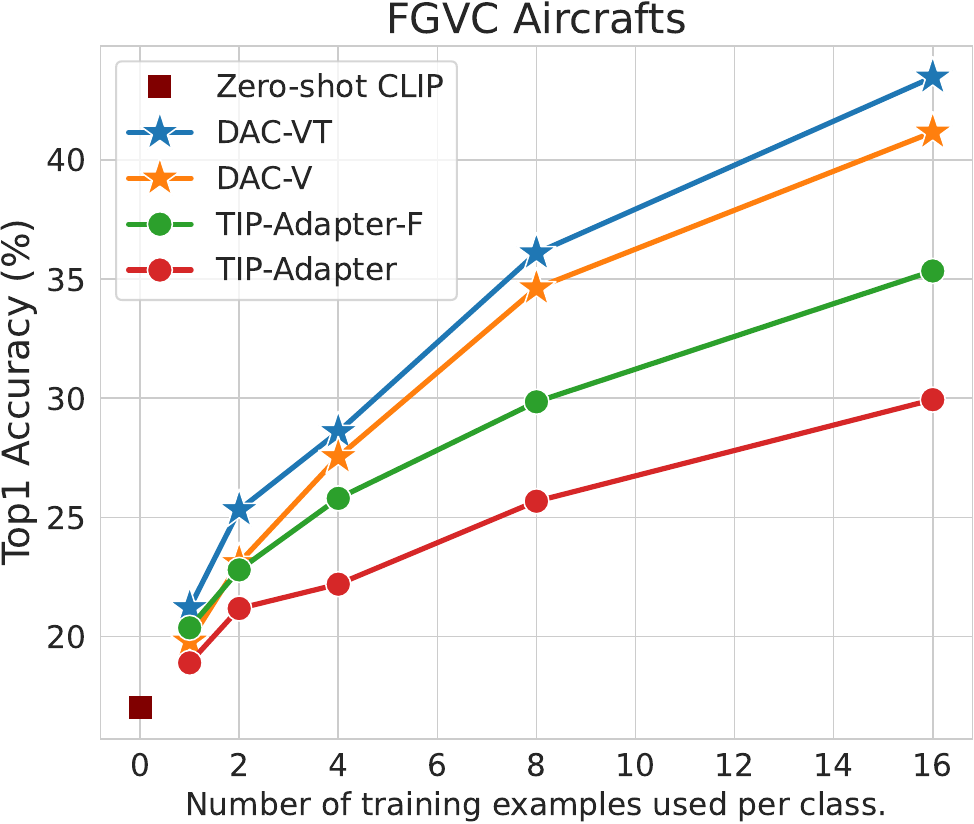}
        \end{subfigure}
        \begin{subfigure}{0.245\textwidth}
            \includegraphics[width=\textwidth]{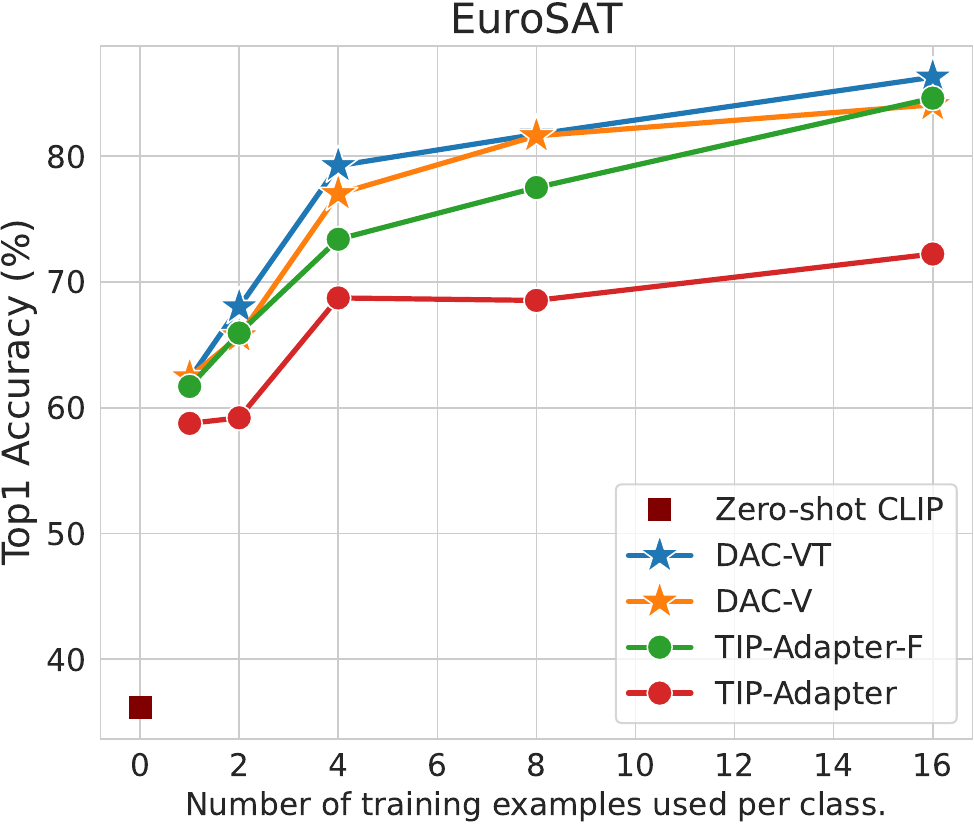}
        \end{subfigure}
        \begin{subfigure}{0.245\textwidth}
            \includegraphics[width=\textwidth]{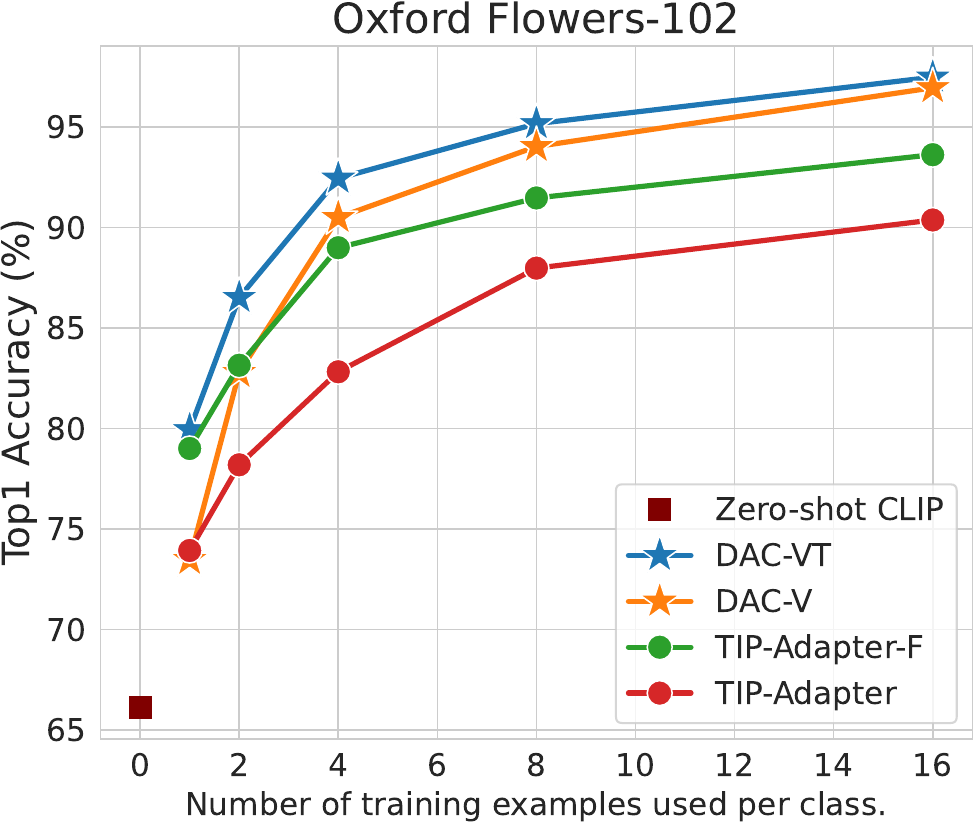}
        \end{subfigure}
        \begin{subfigure}{0.245\textwidth}
            \includegraphics[width=\textwidth]{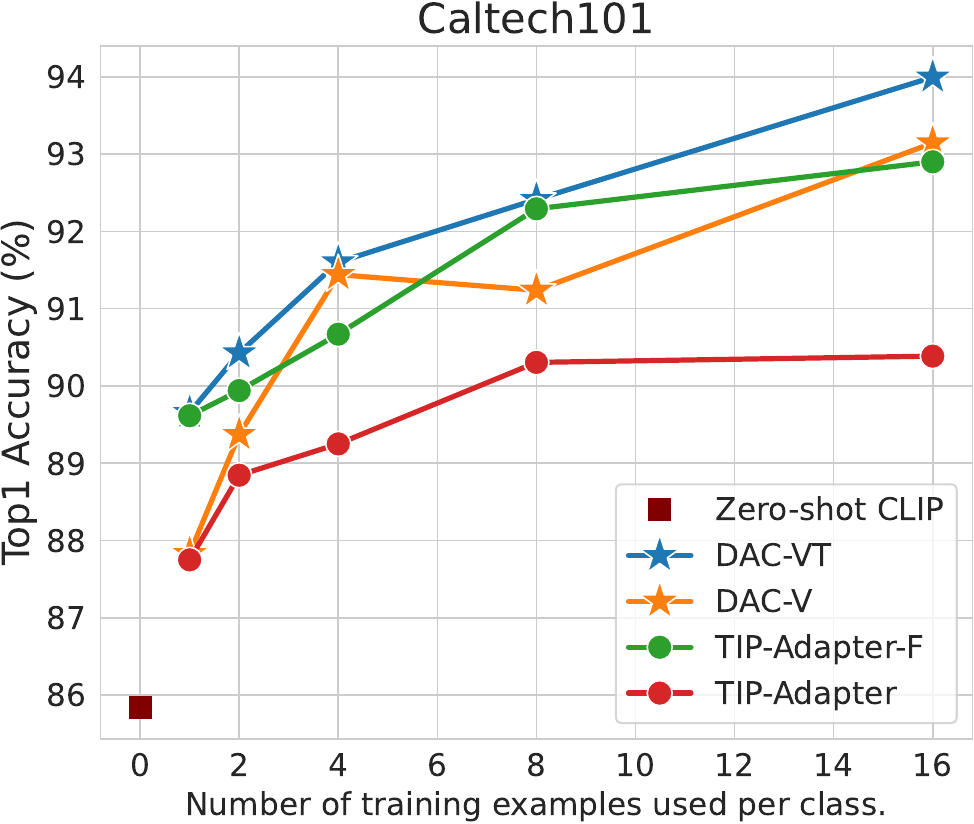}
        \end{subfigure}
        \begin{subfigure}{0.245\textwidth}
            \includegraphics[width=\textwidth]{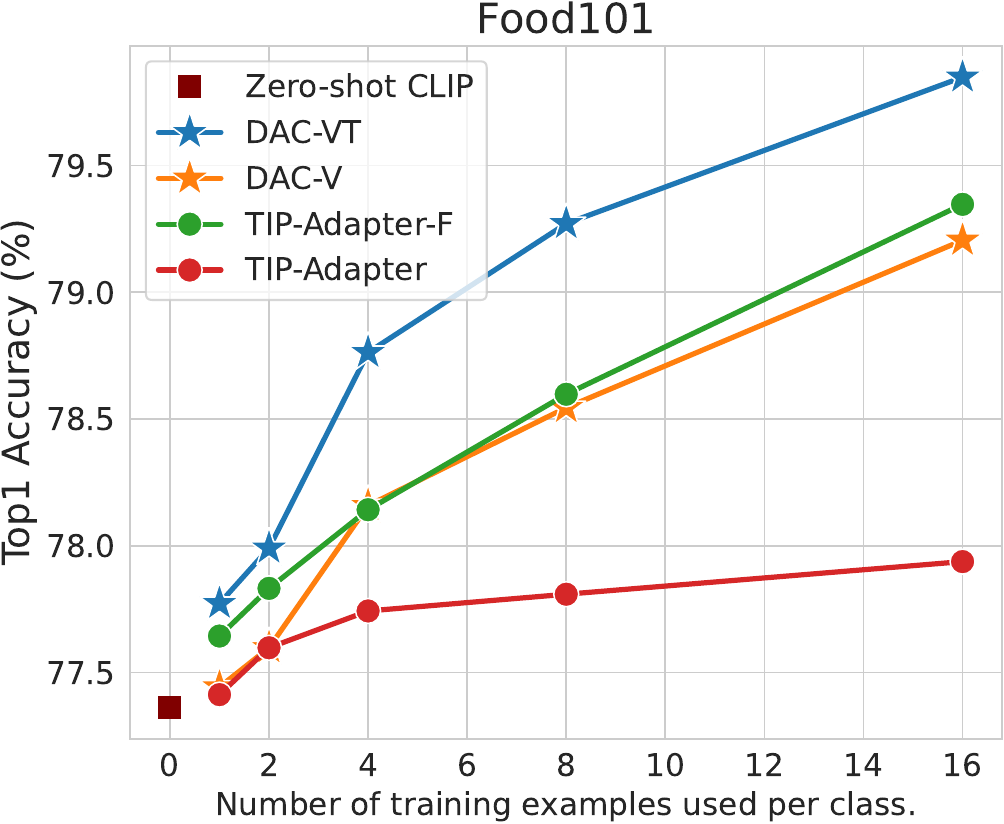}
        \end{subfigure}
        \begin{subfigure}{0.245\textwidth}
            \includegraphics[width=\textwidth]{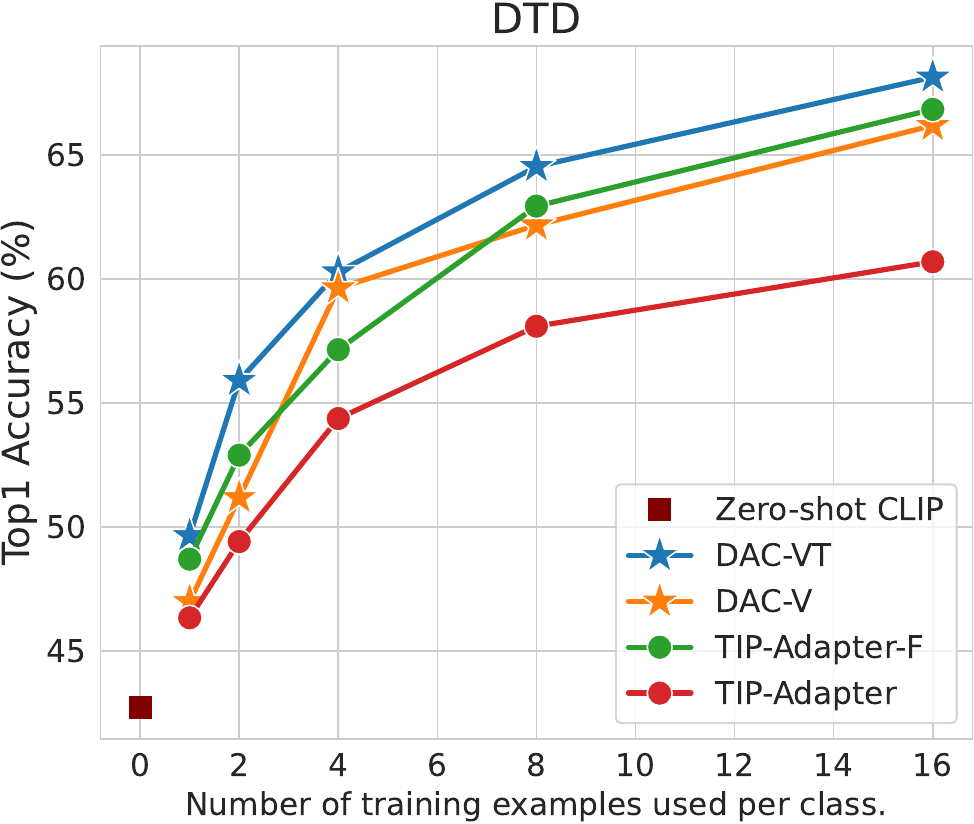}
        \end{subfigure}
        \begin{subfigure}{0.245\textwidth}
            \includegraphics[width=\textwidth]{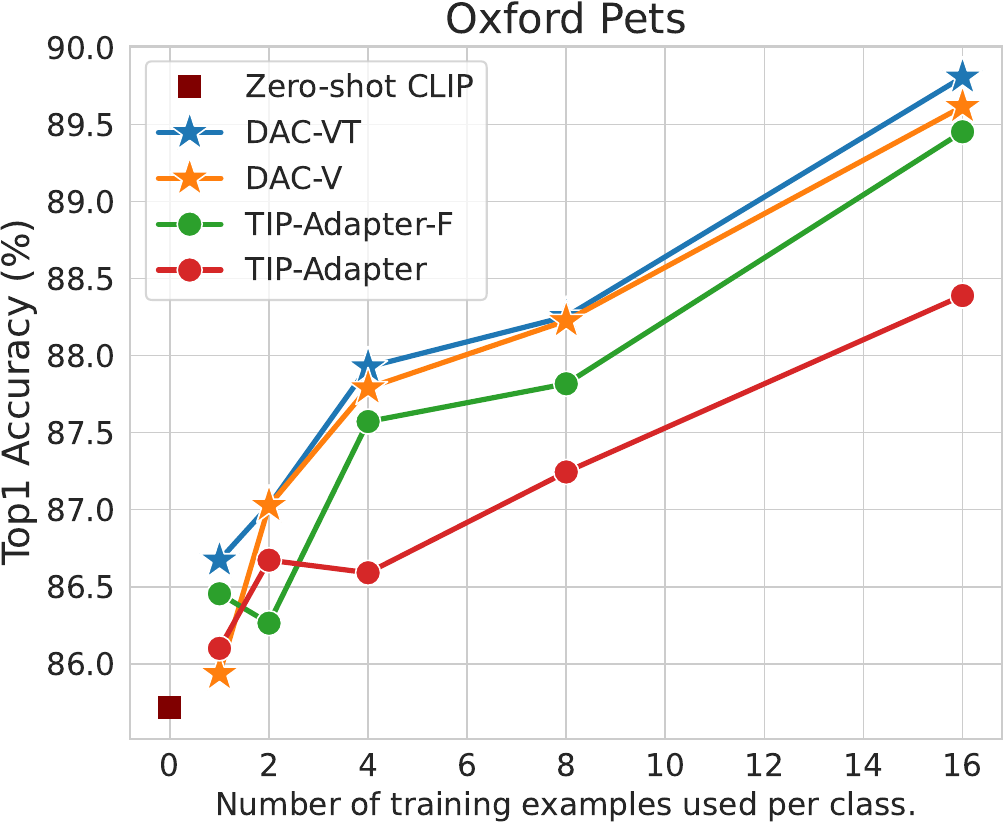}
        \end{subfigure}
        \begin{subfigure}{0.245\textwidth}
            \includegraphics[width=\textwidth]{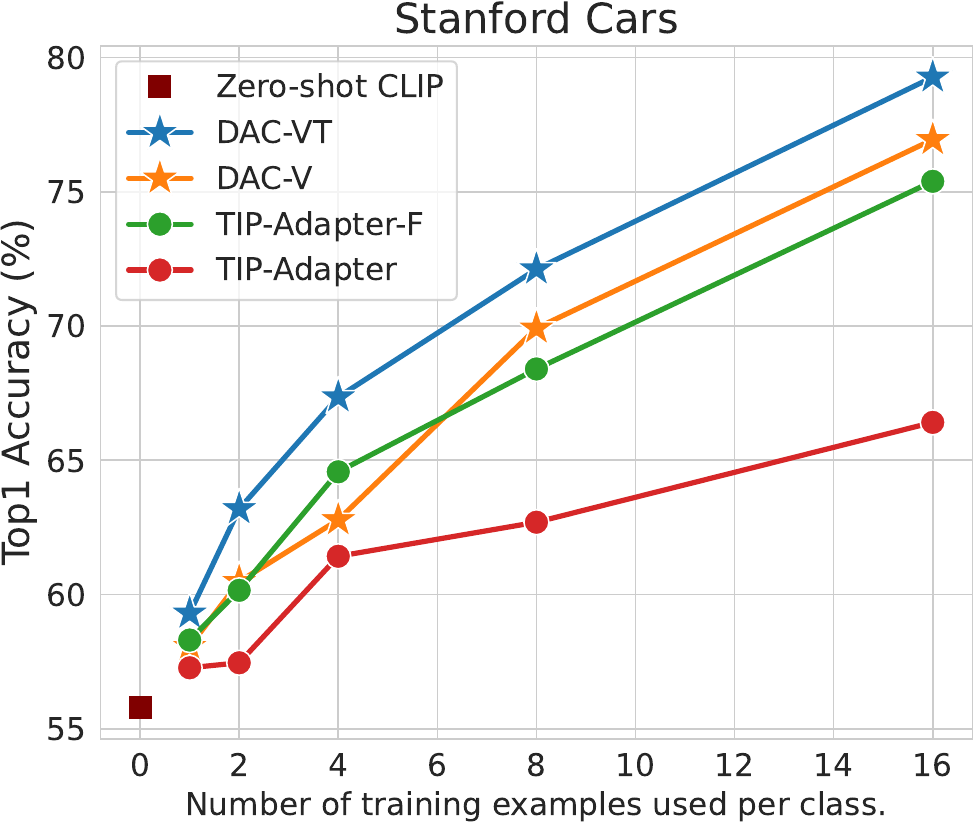}
        \end{subfigure}
        \begin{subfigure}{0.245\textwidth}
            \includegraphics[width=\textwidth]{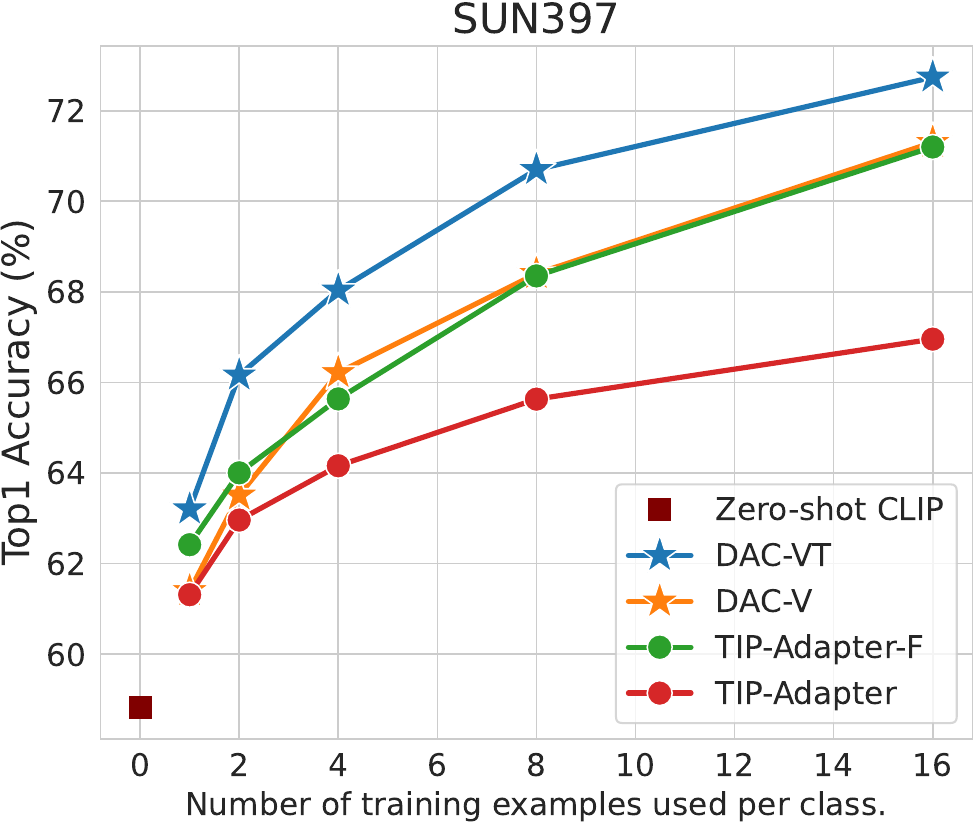}
        \end{subfigure}
        \begin{subfigure}{0.245\textwidth}
            \includegraphics[width=\textwidth]{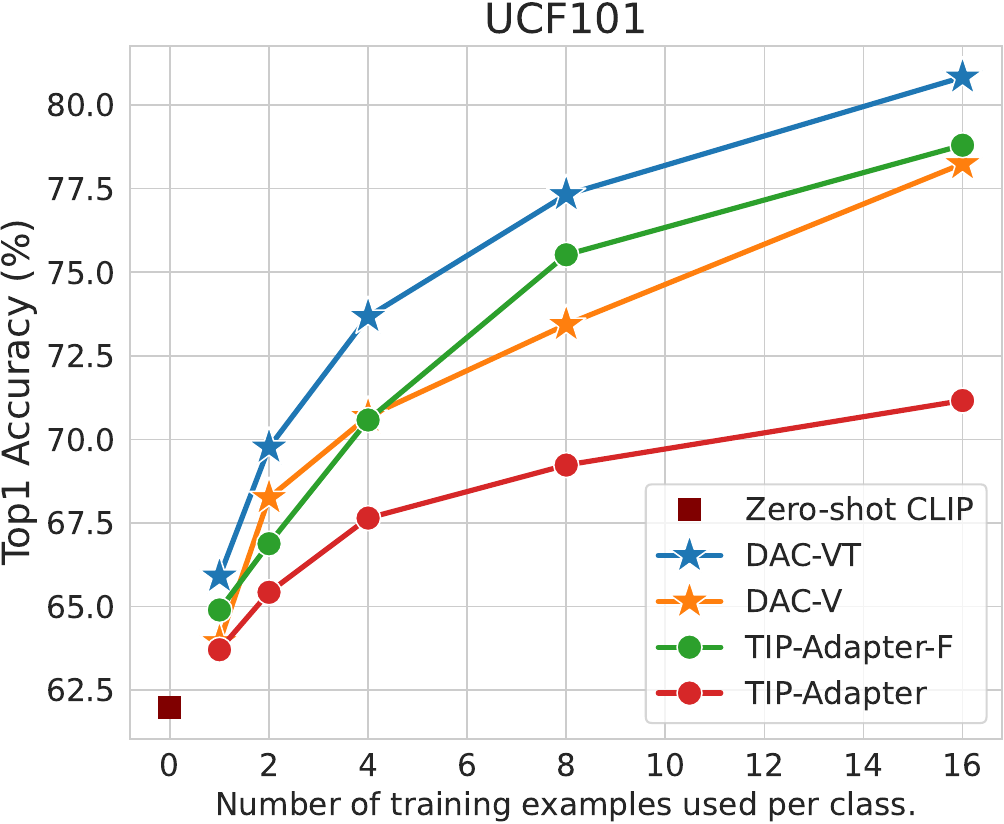}
        \end{subfigure}
    \end{subfigure}
    \caption{{\bf A comparison of top1 accuracy (\%) obtained by different few-shot CLIP adaptation methods over 11 datasets}. Here, the x-axis represents the number of training examples used per class from the target distribution. Our proposed methods, DAC-V which aligns only visual representations of CLIP on the target distribution and DAC-VT which aligns both visual and textual representations, perform comparable or better than the baselines. \textbf{(best viewed in color)}}
    \label{fig:all_comparisons}
\vspace{-1.5em}
\end{figure*}

\noindent
\textbf{Datasets.} For our experiments, we consider ImageNet~\cite{deng2009imagenet}, Caltech101~\cite{fei2004learningcaltech}, FGVCAircraft~\cite{maji2013fineAircraft}, UCF101~\cite{soomro2012ucf101}, EuroSAT~\cite{helber2019eurosat}, Flowers102~\cite{nilsback2008automatedFlower}, StanfordCars~\cite{krause20133dStanfordCars}, DTD~\cite{cimpoi2014describingDtd}, Food101~\cite{bossard2014food}, \mbox{OxfordPets}~\cite{parkhi2012catsOxfordpets}, and, SUN397~\cite{xiao2010sun397}.

\noindent
\textbf{Training and Evaluation Protocol.} We follow the few-shot protocol by \cite{radford2021learning, zhang2021tip} and fine-tune our models using 1, 2, 4, 8, and 16 shots per class, sampled from the training sets. Based on the validation sets, we then select the best fine-tuned adapters and the optimal $\alpha$. Finally, we evaluate on the respective test sets. For ImageNet, like \cite{zhang2021tip}, we report results on the validation sets. Using Adam \cite{kingma2014adam}, we train the visual adapter for 500 epochs with learning rate~0.00003, temperature $\tau$ 0.008 and a batch size equal to the number of classes in the dataset. We set the number of randomly augmented views $M$ to 7, and ablate over this parameter in \cref{sec:ablations}. The training augmentations include random horizontal flips, random cropping, and, resizing to 224$\times$224 pixels. To train the textual adapter, we follow the same data pre-processing protocol, but fine-tune for only 100 epochs with a learning rate of 0.00001. At inference time, we apply CLIP's pre-processing (center cropping and resizing). On a single Nvidia V100 GPU, visual adapter training takes $\sim$1 hour for the 16-shot setting. In contrast, textual adapter training only takes about 30 seconds. Note that for building the visual cache, following \cite{zhang2021tip}, we randomly augment each training image 10 times and use the mean embedding as a cache entry. For a fair comparison, we apply the prompt ensembling of \cite{zhang2021tip} for ImageNet and a single prompt for the other datasets.

\noindent
\textbf{Baselines.} We compare DAC with the strong existing few-shot adaptation methods for CLIP. This includes linear-probe CLIP \cite{radford2021learning}, CoOp \cite{zhou2022learning}, CLIP-Adapter \cite{gao2021clip}, and TIP-Adapter \cite{zhang2021tip}. Note that we do not compare with \cite{xiao2022exploiting, zhang2023prompt} as these recent works leverage language and vision generative models to generate bigger training sets from a few examples. We reproduce the results for TIP-Adapter using their official code\footnote{https://github.com/gaopengcuhk/Tip-Adapter}. For other baselines, we provide officially reported scores for a fair comparison.

\noindent
\textbf{Results and Discussion.} 
In \cref{fig:all_comparisons}, we compare the few-shot classification performances of DAC-V and DAC-VT with Tip-Adapter variants on all the datasets. It can be seen that DAC-V performs comparable to the strong baseline of Tip-Adapter-F. With an increasing number of training shots (\cf 4, 8, and 16-shot), the average performance of DAC-V surpasses that of Tip-Adapter-F by ~0.8\% or (performs better on 6 out of 11 datasets). Note that DAC-V is only optimized to align visual representations in the target domain and no explicit fine-tuning is done to increase its upstream few-shot performance. This clearly demonstrates the benefit of having a strong intra-modal classifier. With further optimization for inter-modal alignment on DAC-V, our proposed DAC-VT method surpasses all baselines by a significant margin. This strong result further illustrates the benefits of aligning both visual and textual domains on target distributions. The results in \cref{fig:all_comparisons} correspond to the ResNet-50 variant of CLIP. In \cref{tab:clip-variants}, we present few-shot adaptation results on ImageNet validation sets using different CLIP backbones. The results indicate a robust performance of DAC-VT across all CLIP variants, significantly outperforming the other baselines.

\begin{table}%
    \centering
    \begin{tabular}{l@{\hskip 0.05in}c@{\hskip 0.05in}c@{\hskip 0.05in}c@{\hskip 0.05in}c@{\hskip 0.05in}c@{\hskip 0.05in}c@{\hskip 0.05in}c}
    \toprule
    Models &\ RN50\ &\ RN101\ &\ V-B/32\ & \ V-B/16\ &\ V-L/14\\
    \midrule
    Zero-shot CLIP &60.33 &62.53 &63.80 &68.73 &75.92\\
	CoOp & 62.95 &66.60 & 66.85  &71.92 &-\\
	CLIP-Adapter & 63.59 & 65.39 & 66.19 &71.13 &-\\
	SgVA-CLIP & 65.70 & 68.51 & 68.26 & 73.30 & - \\
	Tip-Adapter & 62.03 & 64.79 & 65.60 & 70.83 &77.70\\
    Tip-Adapter-F & 65.47 & 68.53 & 68.74 & 73.70 &79.43\\
    \midrule
    \textbf{DAC-V} & 64.89 & 67.38 & 67.77 & 72.98 & 79.62 \\
    \textbf{DAC-VT} & \textbf{66.61} & \textbf{69.37} & \textbf{69.64} & \textbf{74.59} & \textbf{80.20} \\
    \bottomrule
    \end{tabular}
\caption{{\bf 16-shot classification performance of different methods using different CLIP variants on ImageNet}. %
Here, \textit{RN} refers to ResNet and \textit{V} refers to ViT. For Eg., V-B/32 $\rightarrow$ ViT-B/32.}
    \label{tab:clip-variants}
\vspace{-1.5em}
\end{table}

\myparagraph{Distributional Robustness.}
Radford~\etal~\cite{radford2021learning} show that while fine-tuning improves the in-distribution performance, it reduces the overall robustness to shifts in distributions. So far we have observed that improving both visual and textual representations of CLIP in new domains consistently enhances its downstream performance in that domain. However, does it come at the cost of reduced robustness to natural shifts in distributions? In this section, we study the transfer of DAC models trained on ImageNet to four ImageNet variants \ie ImageNet-V2~\cite{recht2019imagenet}, ImageNet-Sketch~\cite{wang2019learning}, ImageNet-A~\cite{hendrycks2021nae} and ImageNet-R~\cite{hendrycks2021many}. In \cref{tab:domain_generalization}, we conduct a cross-dataset evaluation and find that intra-modal alignment (DAC-V) results in better OOD performance, when compared to other methods that aim for inter-modal alignment. We conjecture that alignment of visual features is more robust to distribution shifts than inter-modal alignment \cite{yang2022vision}.
Note that Tip-Adapter-F, optimized for inter-modal alignment, also does not outperform its un-tuned version in OOD setting.

\begin{table}%
\vspace*{0.2cm}
\centering
\small
\begin{tabular}{l@{\hskip 0.05in}c@{\hskip 0.05in}ccccc}
\toprule
\multirow{2}{*} & \textbf{Source} &\multicolumn{4}{c}{\textbf{Target Datasets}} \\
\cmidrule(lr){2-2} \cmidrule(lr){3-6} 
& ImageNet  & -V2 &-A &-R & -Sketch \\
Linear-probe CLIP  & 56.13  & 45.61 &12.71 &34.86 & 19.13\\
CoOp & 62.95  & 54.58 &23.06 &54.96 & 31.04  \\
CoCoOp & 62.81 & 55.72 & 23.32 & 57.74 & 34.48 \\
CALIP-FS & 65.81  & 55.98 & 23.42 & 56.74 & 35.37 \\
Tip-Adapter & 62.03  & 54.56 &23.61 & 60.33 & 35.86  \\
Tip-Adapter-F & 65.47  & 56.79 &20.93 & 58.48 & 34.62  \\
\midrule
DAC-V & 64.89  & 56.56 & \textbf{23.92} & \textbf{60.52} & \textbf{36.27} \\
DAC-VT & \textbf{66.61}  & \textbf{57.68} & 20.92 & 58.68 & 35.33 \\
\bottomrule
\end{tabular}
\caption{{\bf Robustness to Distributional Shifts}. We use CLIP ResNet-50 backbone for all the methods. Here, DAC-V demonstrates better performance than other baselines.}
\label{tab:domain_generalization}
\vspace{-2em}
\end{table}

\begin{figure} %
    \centering

    \begin{subfigure}{1\linewidth}
        \begin{subfigure}{0.49\textwidth}
            \includegraphics[width=\textwidth]{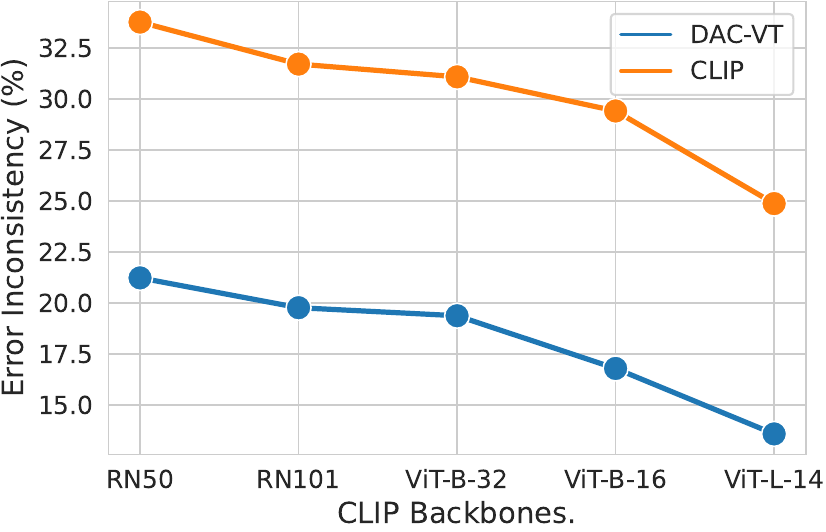}
        \end{subfigure}%
        \begin{subfigure}{0.49\textwidth}
            \includegraphics[width=\textwidth]{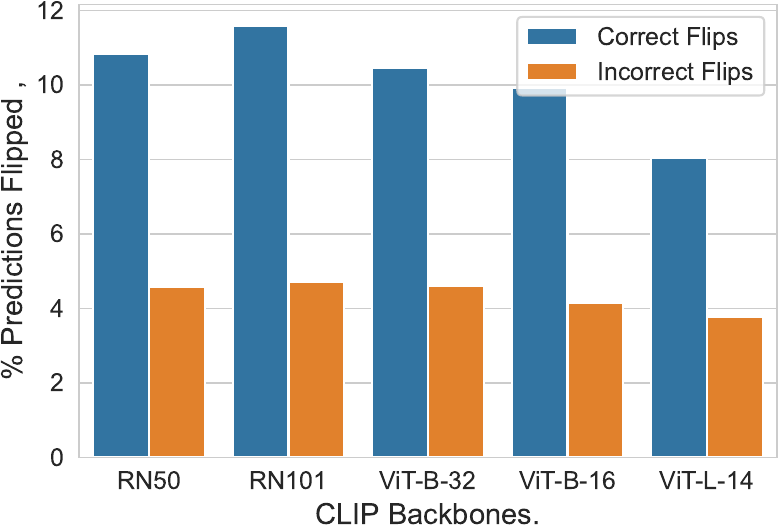}
        \end{subfigure}
    \end{subfigure}
\caption{
{\bf(left)} Error inconsistencies between inter- and intra-modal classifiers using different CLIP backbones on ImageNet. 
DAC-VT significantly reduces this inconsistency.
{\bf(right)} Percentage of correct vs incorrect prediction flips in DAC-VT.}
\label{fig:error_inconsistency}
\vspace{-1.5em}
\end{figure}

\myparagraph{Assaying Inter- and Intra-modal Classifiers in DAC.}
\label{exp:inter_vs_intra}
The results in \cref{fig:all_comparisons} and \cref{tab:clip-variants} empirically verify our main hypothesis that an ensemble of strong intra- and inter-modal classifiers leads to a better overall classifier. However, two questions arise: (1) How much does the visual cache benefit from intra-modal contrastive learning? And, (2) do inter- and intra-modal classifiers make sufficiently uncorrelated mistakes to justify ensembling? \Cref{fig:intra_vs_inter} demonstrates that intra-modal contrastive fine-tuning enhances the discriminative cability of the visual cache, increasing DAC-Cache's performance by 17\% in the 16-shot setting. In contrast, Tip-Adapter-F fine-tuning reduces this performance by implicitly forcing the visual cache to learn the residual information required to improve the upstream (inter-modal) classification, indicating that intra-modal adaptation is indeed beneficial. To analyze whether ensembling inter- and intra-modal classifiers would improve the overall performance, we plot their error inconsistencies in \cref{fig:error_inconsistency}~(left). This reveals that, using pre-trained CLIP features, classifiers make highly uncorrelated mistakes, presenting an opportunity to flip the incorrect predictions via ensembling. DAC-VT reduces this error inconsistency, meaning that the predictions of both classifiers are flipped either correctly or incorrectly.  In \cref{fig:error_inconsistency}~(right) we show that the percentage of correct flips is more than double than that of incorrect flips, indicating that the reduction in error inconsistency is due to correctly flipped predictions in ensembled setting. We refer readers to the ablations in \cref{sec:ablation_err_inconsis} to see a similar behavior of error inconsistencies across different datasets.

\vspace{-0.5em}
\section{Ablations}
\label{sec:ablations}
\begin{figure}[h]
    \centering
    \begin{subfigure}{1\linewidth}
        \begin{subfigure}[t]{0.49\textwidth}
            \includegraphics[width=\textwidth]{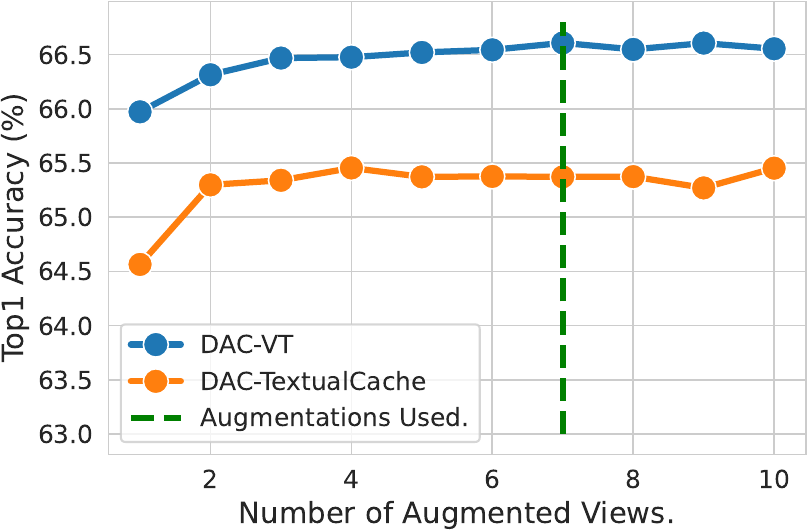}
        \end{subfigure}
        \begin{subfigure}{0.49\textwidth}
            \includegraphics[width=\textwidth]{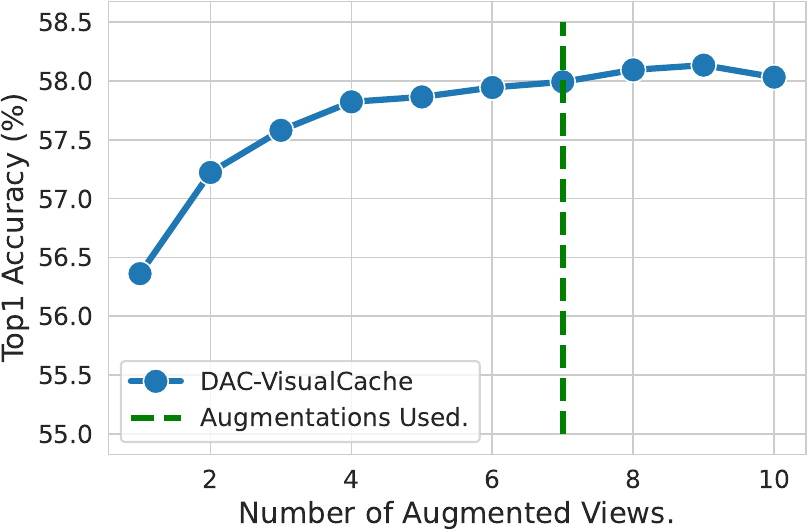}
        \end{subfigure}%
    \end{subfigure}
    \caption{{\bf Effects of varying number of augmented views used to train DAC inter-modal adapter (left) and intra-modal adapter (right)}. The vertical green line indicates the number of augmented views used for all experiments in this paper.
    } 
    \label{fig:augmentations}
\vspace{-1.5em}
\end{figure}

In this section, we ablate over all the components used in the construction of DAC-VT to justify our design choices. In \cref{tab:components-ablation}, we see how the four components of DAC-VT interact on the ImageNet dataset using a CLIP ResNet50 backbone. Our textual adaptation alone provides a significant 5\% boost in performance over zero-shot CLIP's performance (i.e., 60.33\%).
Ensembling it with visual adaptation contributes to an additional 1.3\% gain in accuracy. We also observe that the number of randomly augmentated views $M$ and the weighting parameter $\alpha$ play crucial roles in finding the optimal performance. We further ablate over the number of augmented views in \cref{fig:augmentations}. The accuracy increases monotonically up to $M=7$ augmented views. Therefore, we select $M=7$ for all of our experiments. To select the optimal weighting parameter $\alpha$, we use grid search on the validation sets of each dataset (range [0.1, 10]). Note that \cite{zhang2021tip} used the same strategy to find the optimal residual parameter. For the ImageNet 16-shot classification setting, we empirically find the values of $8.3$ and $3.3$ to be optimal for DAC-V and DAC-VT, respectively. We also ablate over the depth of the visual adapter layer $\linearlayer_{\bm{\theta}}$ and find a single linear layer to be optimal (66.61\% vs 65.58\% of double layered adapter). We provide additional ablations on alpha values and adapter layers in \cref{sec:ablation_app}.

\begin{table}%
    \centering
    \begin{tabular}{cccccc}
    \toprule
    V.A. &\ T.A. &\ $\alpha$\ &\ Augmented Views &\ Top1 (\%)\\
    \midrule
    $\Diamond$ &- &- &- & 41.51\\
    \checkmark &- &- &- & 56.36\\
    \checkmark &- &- &\checkmark & 58.00\\
    \checkmark &$\Diamond$ &- &\checkmark & 62.01\\
    \checkmark &$\Diamond$ &\checkmark &\checkmark & 64.89\\
    -& \checkmark &- &- & 64.56\\
    -& \checkmark &- &\checkmark & 65.37\\
    \checkmark& \checkmark &- &- & 65.33\\
    \checkmark& \checkmark &\checkmark &- & 65.97\\
    \checkmark& \checkmark &- &\checkmark & 66.07\\
    \checkmark& \checkmark &\checkmark &\checkmark & 66.61\\
    \bottomrule
    \end{tabular}
    \caption{{\bf Ablation of DAC's components on ImageNet using ResNet-50 in 16-shot setting}. Here, \mbox{``\textbf{-}''} indicates non-existence of the corresponding feature, \textbf{$\Diamond$} indicates the usage of frozen CLIP features and \textbf{$\checkmark$} indicates their adaptation for DAC-VT. Here V.A and T.A stands for visual and textual adaptation respectively.}
    \label{tab:components-ablation}
\vspace{-1.5em}
\end{table}

\noindent
\textbf{Ablating Alternative Ensembling Choices.}
In addition to the proposed DAC-VT framework, we also looked at other alternatives. The following experiments consider 16-shot classification on ImageNet using CLIP ResNet-50. 

\noindent
\textbf{End-to-end Visual and Textual Domains Adaptation.} We tried to adapt both visual and textual representations together in an end-to-end fashion. The strategy is similar to CLIP-Adapter \cite{gao2021clip}, however, we use a linear layer for visual features adaption as in DAC-V and adapted textual features as described in \cref{sec:text_alignment}. The resulting model obtained 64.22\% (vs 66.61\% of DAC-VT). Noticeably, the intra-modal classification deteriorated similar to Tip-Adapter \ie 28.44\% (vs 41.51\% of CLIP). 

\noindent
\textbf{Building Visual Cache with Class Prototypes.} Instead of retaining all image embeddings in the visual cache, we can reduce them to class prototypes \cite{snell2017prototypical} \ie averaging image embeddings corresponding to all classes. This approach in DAC-VT's style ensembling obtained 66.06\% while it obtained 65.83\% top1 accuracy in an end-to-end setting.

\noindent
\textbf{Cross-entropy Loss for Training Visual Adapter.}
The visual adapter can also be trained via cross-entropy instead of a contrastive objective.
We ran an experiment where apart from the visual adapter's training objective, all settings remained the same. 
The cross-entropy method gets 65.62\% accuracy, while the contrastive objective achieves 66.61\%.

\vspace{-0.5em}
\section{Conclusion}
We presented a sample-efficient framework, DAC, for adapting CLIP to downstream classification tasks. Using only a few labeled examples from a target distribution, DAC boosts the overall classification performance by improving both intra- and inter-modal representations of CLIP. Extensive experiments on 11 widely used image classification benchmarks show that DAC outperforms the competitive baselines while maintaining robustness to natural shifts in distributions.
The performance improvements come at negligible additional computational cost during inference,
as our framework requires only a linear layer for adaptation.
Although the inference cost of DAC remains low, the two-stage adaptation increases the computational overhead for fine-tuning in comparison to the competitive baselines.
We posit that there is room for further improving the ensembling of intra- and inter-modal classifiers, as both classifiers continue to exhibit uncorrelated errors \cf \cref{fig:error_inconsistency}.

{\small
\balance
\bibliographystyle{ieee_fullname}
\bibliography{arxiv}

\begin{thebibliography}{10}\itemsep=-1pt

\bibitem{bengio2013representation}
Yoshua Bengio, Aaron Courville, and Pascal Vincent.
\newblock Representation learning: {A} review and new perspectives.
\newblock {\em IEEE T. Pattern Anal. Mach. Intell.}, 35(8):1798--1828, Aug.
  2013.

\bibitem{beyer2020we}
Lucas Beyer, Olivier~J. H{\'{e}}naff, Alexander Kolesnikov, Xiaohua Zhai, and
  A{\"{a}}ron van~den Oord.
\newblock Are we done with {ImageNet}?
\newblock {\em arXiv preprint arXiv:2006.07159}, 2020.

\bibitem{bossard2014food}
Lukas Bossard, Matthieu Guillaumin, and Luc~Van Gool.
\newblock {Food-101} -- {M}ining discriminative components with random forests.
\newblock In {\em ECCV}, volume~6, pages 446--461, 2014.

\bibitem{chen2020simple}
Ting Chen, Simon Kornblith, Mohammad Norouzi, and Geoffrey~E. Hinton.
\newblock A simple framework for contrastive learning of visual
  representations.
\newblock In {\em ICML}, pages 1597--1607, 2020.

\bibitem{cimpoi2014describingDtd}
Mircea Cimpoi, Subhransu Maji, Iasonas Kokkinos, Sammy Mohamed, and Andrea
  Vedaldi.
\newblock Describing textures in the wild.
\newblock In {\em CVPR}, pages 3606--3613, 2014.

\bibitem{deng2009imagenet}
Jia Deng, Wei Dong, Richard Socher, Li{-}Jia Li, Kai Li, and Li Fei{-}Fei.
\newblock {ImageNet}: {A} large-scale hierarchical image database.
\newblock In {\em CVPR}, pages 248--255, 2009.

\bibitem{duan2022multi}
Jiali Duan, Liqun Chen, Son Tran, Jinyu Yang, Yi Xu, Belinda Zeng, and Trishul
  Chilimbi.
\newblock Multi-modal alignment using representation codebook.
\newblock In {\em CVPR}, pages 15651--15660, 2022.

\bibitem{fang2022data}
Alex Fang, Gabriel Ilharco, Mitchell Wortsman, Yuhao Wan, Vaishaal Shankar,
  Achal Dave, and Ludwig Schmidt.
\newblock Data determines distributional robustness in contrastive language
  image pre-training {(CLIP)}.
\newblock In {\em ICML}, pages 6216--6234, 2022.

\bibitem{fei2004learningcaltech}
Li Fei-Fei, Rob Fergus, and Pietro Perona.
\newblock Learning generative visual models from few training examples: {A}n
  incremental {B}ayesian approach tested on 101 object categories.
\newblock {\em Comput. Vis. Image Und.}, 106(1):59--70, Apr. 2007.

\bibitem{gao2021clip}
Peng Gao, Shijie Geng, Renrui Zhang, Teli Ma, Rongyao Fang, Yongfeng Zhang,
  Hongsheng Li, and Yu Qiao.
\newblock {CLIP-Adapter}: {B}etter vision-language models with feature
  adapters.
\newblock {\em arXiv preprint arXiv:2110.04544}, 2021.

\bibitem{gondal2021function}
Muhammad~Waleed Gondal, Shruti Joshi, Nasim Rahaman, Stefan Bauer, Manuel
  Wuthrich, and Bernhard Sch{\"o}lkopf.
\newblock Function contrastive learning of transferable meta-representations.
\newblock In {\em ICML}, pages 3755--3765, 2021.

\bibitem{gontijo2021no}
Raphael Gontijo{-}Lopes, Yann~N. Dauphin, and Ekin~D. Cubuk.
\newblock No one representation to rule them all: {O}verlapping features of
  training methods.
\newblock In {\em ICLR}, 2022.

\bibitem{guo2022calip}
Ziyu Guo, Renrui Zhang, Longtian Qiu, Xianzheng Ma, Xupeng Miao, Xuming He, and
  Bin Cui.
\newblock {CALIP}: {Z}ero-shot enhancement of {CLIP} with parameter-free
  attention.
\newblock In {\em AAAI}, pages 746--754, 2023.

\bibitem{gutmann2010noise}
Michael Gutmann and Aapo Hyv{\"a}rinen.
\newblock Noise-contrastive estimation: {A} new estimation principle for
  unnormalized statistical models.
\newblock In {\em AISTATS}, pages 297--304, 2010.

\bibitem{he2020momentum}
Kaiming He, Haoqi Fan, Yuxin Wu, Saining Xie, and Ross~B. Girshick.
\newblock Momentum contrast for unsupervised visual representation learning.
\newblock In {\em CVPR}, pages 9729--9738, 2020.

\bibitem{helber2019eurosat}
Patrick Helber, Benjamin Bischke, Andreas Dengel, and Damian Borth.
\newblock {EuroSAT}: {A} novel dataset and deep learning benchmark for land use
  and land cover classification.
\newblock {\em {IEEE} J. Sel. Top. Appl. Earth Obs. Remote. Sens.},
  12(7):2217--2226, July 2019.

\bibitem{hendrycks2021many}
Dan Hendrycks, Steven Basart, Norman Mu, Saurav Kadavath, Frank Wang, Evan
  Dorundo, Rahul Desai, Tyler Zhu, Samyak Parajuli, Mike Guo, Dawn Song, Jacob
  Steinhardt, and Justin Gilmer.
\newblock The many faces of robustness: {A} critical analysis of
  out-of-distribution generalization.
\newblock In {\em ICCV}, pages 8320--8329, 2021.

\bibitem{hendrycks2021nae}
Dan Hendrycks, Kevin Zhao, Steven Basart, Jacob Steinhardt, and Dawn Song.
\newblock Natural adversarial examples.
\newblock In {\em CVPR}, pages 15262--15271, 2021.

\bibitem{houlsby2019parameter}
Neil Houlsby, Andrei Giurgiu, Stanis\l{}aw Jastrz\k{e}bski, Bruna Morrone,
  Quentin de Laroussilhe, Andrea Gesmundo, Mona Attariyan, and Sylvain Gelly.
\newblock Parameter-efficient transfer learning for {NLP}.
\newblock In {\em ICML}, pages 2790--2799, 2019.

\bibitem{jia2021scaling}
Chao Jia, Yinfei Yang, Ye Xia, Yi{-}Ting Chen, Zarana Parekh, Hieu Pham,
  Quoc~V. Le, Yun{-}Hsuan Sung, Zhen Li, and Tom Duerig.
\newblock Scaling up visual and vision-language representation learning with
  noisy text supervision.
\newblock In {\em ICML}, pages 4904--4916, 2021.

\bibitem{kingma2014adam}
Diederik~P. Kingma and Jimmy~Lei Ba.
\newblock Adam: {A} method for stochastic optimization.
\newblock In {\em ICLR}, 2015.

\bibitem{kolesnikov2020big}
Alexander Kolesnikov, Lucas Beyer, Xiaohua Zhai, Joan Puigcerver, Jessica Yung,
  Sylvain Gelly, and Neil Houlsby.
\newblock Big transfer ({BiT}): {G}eneral visual representation learning.
\newblock In {\em ECCV}, volume~5, pages 491--507, 2020.

\bibitem{krause20133dStanfordCars}
Jonathan Krause, Michael Stark, Jia Deng, and Li Fei{-}Fei.
\newblock {3D} object representations for fine-grained categorization.
\newblock In {\em ICCVW}, pages 554--561, 2013.

\bibitem{liang2022mind}
Victor~Weixin Liang, Yuhui Zhang, Yongchan Kwon, Serena Yeung, and James~Y Zou.
\newblock Mind the gap: {U}nderstanding the modality gap in multi-modal
  contrastive representation learning.
\newblock In {\em NeurIPS}, pages 17612--17625, 2022.

\bibitem{ma2022understanding}
Chengcheng Ma, Yang Liu, Jiankang Deng, Lingxi Xie, Weiming Dong, and
  Changsheng Xu.
\newblock Understanding and mitigating overfitting in prompt tuning for
  vision-language models.
\newblock {\em arXiv preprint arXiv:2211.02219}, 2022.

\bibitem{maji2013fineAircraft}
Subhransu Maji, Esa Rahtu, Juho Kannala, Matthew~B. Blaschko, and Andrea
  Vedaldi.
\newblock Fine-grained visual classification of aircraft.
\newblock {\em arXiv preprint arXiv:1306.5151}, 2013.

\bibitem{neyshabur2020being}
Behnam Neyshabur, Hanie Sedghi, and Chiyuan Zhang.
\newblock What is being transferred in transfer learning?
\newblock In {\em NeurIPS}, pages 512--523, 2020.

\bibitem{nilsback2008automatedFlower}
Maria{-}Elena Nilsback and Andrew Zisserman.
\newblock Automated flower classification over a large number of classes.
\newblock In {\em ICVGIP}, pages 722--729, 2008.

\bibitem{opitz1999popular}
David~W. Opitz and Richard Maclin.
\newblock Popular ensemble methods: {A}n empirical study.
\newblock {\em J. Artif. Intell. Res.}, 11:169--198, Aug. 1999.

\bibitem{parkhi2012catsOxfordpets}
Omkar~M. Parkhi, Andrea Vedaldi, Andrew Zisserman, and C.~V. Jawahar.
\newblock Cats and dogs.
\newblock In {\em CVPR}, pages 3498--3505, 2012.

\bibitem{peng2022sgva}
Fang Peng, Xiaoshan Yang, Linhui Xiao, Yaowei Wang, and Changsheng Xu.
\newblock {SgVA-CLIP}: {S}emantic-guided visual adapting of vision-language
  models for few-shot image classification.
\newblock {\em arXiv preprint arXiv:2211.16191}, 2022.

\bibitem{pham2021combined}
Hieu Pham, Zihang Dai, Golnaz Ghiasi, Kenji Kawaguchi, Hanxiao Liu, Adams~Wei
  Yu, Jiahui Yu, Yi{-}Ting Chen, Minh{-}Thang Luong, Yonghui Wu, Mingxing Tan,
  and Quoc~V. Le.
\newblock Combined scaling for zero-shot transfer learning.
\newblock {\em Neurocomputing}, 555:126658, Oct. 2023.

\bibitem{radford2021learning}
Alec Radford, Jong~Wook Kim, Chris Hallacy, Aditya Ramesh, Gabriel Goh,
  Sandhini Agarwal, Girish Sastry, Amanda Askell, Pamela Mishkin, Jack Clark,
  Gretchen Krueger, and Ilya Sutskever.
\newblock Learning transferable visual models from natural language
  supervision.
\newblock In {\em ICML}, pages 8748--8763, 2021.

\bibitem{recht2019imagenet}
Benjamin Recht, Rebecca Roelofs, Ludwig Schmidt, and Vaishaal Shankar.
\newblock Do {ImageNet} classifiers generalize to {ImageNet}?
\newblock In {\em ICML}, pages 5389--5400, 2019.

\bibitem{ren2022rethinking}
Shuhuai Ren, Lei Li, Xuancheng Ren, Guangxiang Zhao, and Xu Sun.
\newblock Delving into the openness of {CLIP}.
\newblock In {\em ACL}, pages 9587--9606, 2023.

\bibitem{shu2022test}
Manli Shu, Weili Nie, De{-}An Huang, Zhiding Yu, Tom Goldstein, Anima
  Anandkumar, and Chaowei Xiao.
\newblock Test-time prompt tuning for zero-shot generalization in
  vision-language models.
\newblock In {\em NeurIPS}, pages 14274--14289, 2022.

\bibitem{snell2017prototypical}
Jake Snell, Kevin Swersky, and Richard~S. Zemel.
\newblock Prototypical networks for few-shot learning.
\newblock In {\em NIPS}, pages 4077--4087, 2017.

\bibitem{soomro2012ucf101}
Khurram Soomro, Amir~Roshan Zamir, and Mubarak Shah.
\newblock {UCF101}: {A} dataset of 101 human actions classes from videos in the
  wild.
\newblock {\em arXiv preprint arXiv:1212.0402}, 2012.

\bibitem{tian2020contrastive}
Yonglong Tian, Dilip Krishnan, and Phillip Isola.
\newblock Contrastive multiview coding.
\newblock In {\em ECCV}, volume~11, pages 776--794, 2020.

\bibitem{tsipras2020imagenet}
Dimitris Tsipras, Shibani Santurkar, Logan Engstrom, Andrew Ilyas, and
  Aleksander Madry.
\newblock From {ImageNet} to image classification: {C}ontextualizing progress
  on benchmarks.
\newblock In {\em ICML}, pages 9625--9635, 2020.

\bibitem{udandarao2022sus}
Vishaal Udandarao, Ankush Gupta, and Samuel Albanie.
\newblock {SuS-X}: {T}raining-free name-only transfer of vision-language
  models.
\newblock {\em arXiv preprint arXiv:2211.16198}, 2022.

\bibitem{oord2018representation}
A{\"{a}}ron van~den Oord, Yazhe Li, and Oriol Vinyals.
\newblock Representation learning with contrastive predictive coding.
\newblock {\em arXiv preprint arXiv:1807.03748}, 2018.

\bibitem{wang2019learning}
Haohan Wang, Songwei Ge, Zachary~C. Lipton, and Eric~P. Xing.
\newblock Learning robust global representations by penalizing local predictive
  power.
\newblock In {\em NeurIPS}, pages 10506--10518, 2019.

\bibitem{wen2020batchensemble}
Yeming Wen, Dustin Tran, and Jimmy Ba.
\newblock {BatchEnsemble}: {A}n alternative approach to efficient ensemble and
  lifelong learning.
\newblock In {\em ICLR}, 2020.

\bibitem{wortsman2022robust}
Mitchell Wortsman, Gabriel Ilharco, Jong~Wook Kim, Mike Li, Simon Kornblith,
  Rebecca Roelofs, Raphael~Gontijo Lopes, Hannaneh Hajishirzi, Ali Farhadi,
  Hongseok Namkoong, and Ludwig Schmidt.
\newblock Robust fine-tuning of zero-shot models.
\newblock In {\em CVPR}, pages 7959--7971, 2022.

\bibitem{xiao2010sun397}
Jianxiong Xiao, James Hays, Krista~A. Ehinger, Aude Oliva, and Antonio
  Torralba.
\newblock {SUN} database: {L}arge-scale scene recognition from abbey to zoo.
\newblock In {\em CVPR}, pages 3485--3492, 2010.

\bibitem{xiao2022exploiting}
Taihong Xiao, Zirui Wang, Liangliang Cao, Jiahui Yu, Shengyang Dai, and
  Ming{-}Hsuan Yang.
\newblock Exploiting category names for few-shot classification with
  vision-language models.
\newblock {\em arXiv preprint arXiv:2211.16594}, 2022.

\bibitem{yang2022vision}
Jinyu Yang, Jiali Duan, Son Tran, Yi Xu, Sampath Chanda, Liqun Chen, Belinda
  Zeng, Trishul Chilimbi, and Junzhou Huang.
\newblock Vision-language pre-training with triple contrastive learning.
\newblock In {\em CVPR}, pages 15650--15659, 2022.

\bibitem{yao2021cpt}
Yuan Yao, Ao Zhang, Zhengyan Zhang, Zhiyuan Liu, Tat{-}Seng Chua, and Maosong
  Sun.
\newblock {CPT}: {C}olorful prompt tuning for pre-trained vision-language
  models.
\newblock {\em arXiv preprint arXiv:2109.11797}, 2021.

\bibitem{yuan2021florence}
Lu Yuan, Dongdong Chen, Yi{-}Ling Chen, Noel Codella, Xiyang Dai, Jianfeng Gao,
  Houdong Hu, Xuedong Huang, Boxin Li, Chunyuan Li, Ce Liu, Mengchen Liu,
  Zicheng Liu, Yumao Lu, Yu Shi, Lijuan Wang, Jianfeng Wang, Bin Xiao, Zhen
  Xiao, Jianwei Yang, Michael Zeng, Luowei Zhou, and Pengchuan Zhang.
\newblock Florence: {A} new foundation model for computer vision.
\newblock {\em arXiv preprint arXiv:2111.11432}, 2021.

\bibitem{zhang2023prompt}
Renrui Zhang, Xiangfei Hu, Bohao Li, Siyuan Huang, Hanqiu Deng, Yu Qiao, Peng
  Gao, and Hongsheng Li.
\newblock Prompt, generate, then cache: {C}ascade of foundation models makes
  strong few-shot learners.
\newblock In {\em CVPR}, pages 15211--15222, 2023.

\bibitem{zhang2021vt}
Renrui Zhang, Longtian Qiu, Wei Zhang, and Ziyao Zeng.
\newblock {VT-CLIP}: {E}nhancing vision-language models with visual-guided
  texts.
\newblock {\em arXiv preprint arXiv:2112.02399}, 2021.

\bibitem{zhang2021tip}
Renrui Zhang, Wei Zhang, Rongyao Fang, Peng Gao, Kunchang Li, Jifeng Dai, Yu
  Qiao, and Hongsheng Li.
\newblock {Tip-Adapter}: {T}raining-free adaption of {CLIP} for few-shot
  classification.
\newblock In {\em ECCV}, volume~35, pages 493--510, 2022.

\bibitem{zhou2022conditional}
Kaiyang Zhou, Jingkang Yang, Chen~Change Loy, and Ziwei Liu.
\newblock Conditional prompt learning for vision-language models.
\newblock In {\em CVPR}, pages 16795--16804, 2022.

\bibitem{zhou2022learning}
Kaiyang Zhou, Jingkang Yang, Chen~Change Loy, and Ziwei Liu.
\newblock Learning to prompt for vision-language models.
\newblock {\em Int. J. Comput. Vis.}, 130(9):2337--2348, July 2022.

\end{thebibliography}
}

\clearpage
\appendix
\section{Ablations}
\label{sec:ablation_app}
\paragraph{Architecture for Visual Adapter Layer.}
In the main paper, we use a linear layer for adapting visual features. 
To ablate the varying depth of $\linearlayer_{\bm{\theta}}$ , we increase the number of layers (with ReLU activations in between). 
\Cref{tab:h_struct} shows the results on the ImageNet validation set using CLIP RN-50 on 16-shot classification.

\begin{figure}[b]
    \centering
    \begin{subfigure}[t]{1.0\linewidth}
        \includegraphics[width=\textwidth]{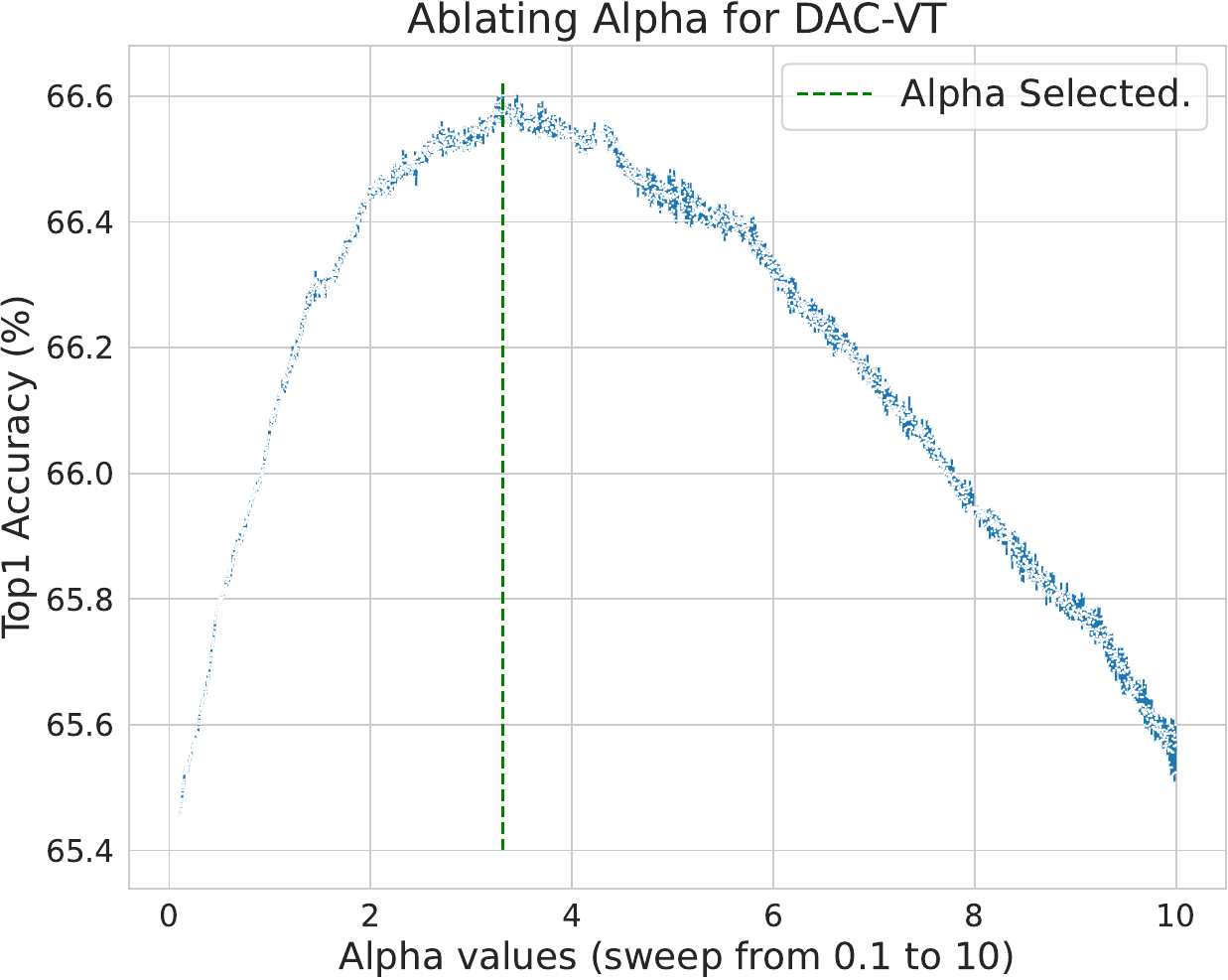}
    \end{subfigure}
    \caption{Ablating the $\alpha$ parameter for ImageNet using CLIP ResNet50.} 
    \label{fig:alpha_ablation}
\end{figure}

\paragraph{Weighting Parameter $\alpha$ for different datasets.}
DAC improves the classification capability of both inter-modal and intra-modal classifiers. We use a scalar $\alpha$ to balance the contributions of each classifier towards the final accuracy. The value for $\alpha$ is selected based on the performance on the validation sets. A similar strategy was employed by Tip-Adapter \cite{zhang2021tip}. However, in Tip-Adapter it is used to determine how much residual information should flow from intra-modal classifier to update the inter-modal predictions. To find the optimal value, we perform a grid search with a step size of 0.01, a search range in [0.1, 10], and the number of search steps being 10000. In this section, we present different values of $\alpha$ used to compute the final test performances of each dataset. In \cref{fig:alpha_ablation}, we show how varying $\alpha$ influences the performance on 16-shot ImageNet classification.  \Cref{tab:dacvt_alphas} lists our optimal values for $\alpha$ for all datasets (both DAC-V and DAC-VT). Since $\alpha$ is multiplied with the intra-modal logits, it can be seen that DAC-V consumes more information from the intra-modal classifier. Remember that in DAC-V, we only optimize the visual representations of CLIP without optimizing it for the upstream few-shot classification task. This further highlights the benefits of having better intra-modal representations in few-shot adaptation setting.

\begin{table}[b] %
\centering
\small
\begin{tabular}{l@{\hskip 0.05in}c@{\hskip 0.05in}c}
\toprule
Structure of $\linearlayer_{\bm{\theta}}$ & DAC-V & DAC-VT \\
\midrule
Linear Layer & 64.89  & 66.61 \\
2 Layer MLP & 64.452  & 65.582 \\
3 Layer MLP & 64.08  & 65.274 \\
4 Layer MLP & 64.01  & 65.064 \\
\bottomrule
\end{tabular}
\caption{Ablating structure of adapter layer $\linearlayer_{\bm{\theta}}$}
\label{tab:h_struct}
\end{table}

\begin{table}[b]
    \centering
    \begin{tabular}{ccc}
    \toprule  
    Datasets & DAC-V $\alpha$  & DAC-VT $\alpha$ \\
    \midrule
    UCF-101 &3.78& 1.16 \\
    Caltech101 &2.40& 1.33 \\
    ImageNet &8.32& 3.31  \\
    SUN397 &5.95& 1.39 \\
    FGVCAircraft & 8.2& 6.91 \\
    StanfordCars &6.50& 2.42  \\
    Flowers102 &8.17& 3.43 \\
    Food101 &1.17& 1.05  \\
    OxfordPets &1.07& 0.73  \\
    DTD & 3.05 & 1.11  \\
    EuroSAT &5.17& 0.76 \\
    \bottomrule
    \end{tabular}
    \caption{Details of $\alpha$ used to weigh intra and inter-modal classifiers for different datasets in DAC-V and DAC-VT. In DAC-V the contribution from intra-modal features is weighted more which indicates that the adapted visual cache contains reliable information to update CLIP's inter-modal knowledge.} 
    \label{tab:dacvt_alphas}
\end{table}

\section{Detailed Analysis on Error Inconsistencies} 
\label{sec:ablation_err_inconsis}
\begin{figure}[b]
    \centering
    \begin{subfigure}[t]{1.0\linewidth}
        \includegraphics[width=\textwidth]{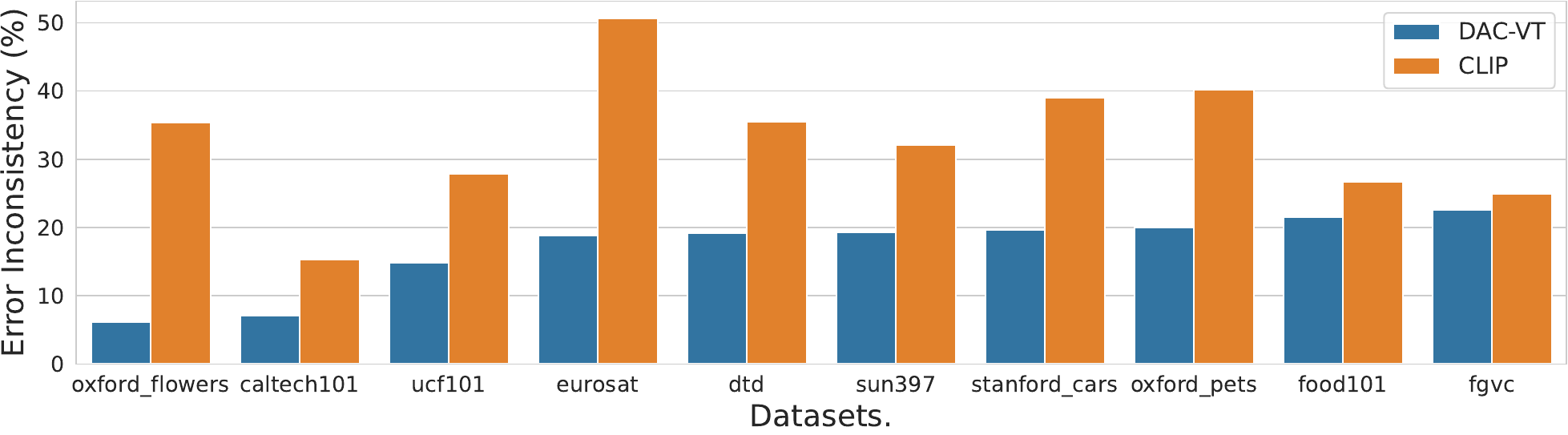}
    \end{subfigure}
    \caption{{\bf Comparative analysis of error inconsistencies between intra-modal and inter-modal classifiers of CLIP and DAC-VT on 10 different datasets} (sorted by DAC-VT's performance). We observe that DAC-VT significantly reduces the error inconsistencies, however, the performance gap reduces on certain datasets such Food101 and FGVCAircrafts.} 
    \label{fig:error_inconsistency_all}
\end{figure}

We analyze the error inconsistencies observed across various datasets in \cref{fig:error_inconsistency_all}. This plot complements our analysis in \cref{sec:experiments} about the role of inter- and intra-modal classifiers in an ensembled setting, and further illustrates how DAC-VT reduces inconsistencies between intra and inter-modal classifiers. The consensus between the DAC-VT's sub-classifiers is higher for some datasets (\eg, {Flowers102}, {Caltech101}), however, the inconsistencies for certain datasets (\eg, FGVCAircraft) are still high.

\section{A Case for Aligning Textual Representations in Target Domain}
\label{sec:aligning_textual_downstream_tasks}
\begin{figure}[b]
    \centering
    \begin{subfigure}[t]{0.48\linewidth}
        \includegraphics[width=\textwidth]{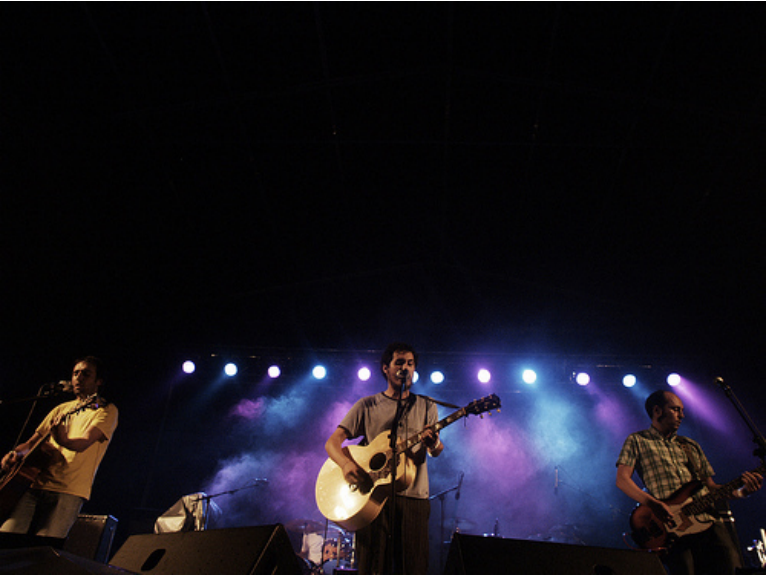}
        \caption{stage}
    \end{subfigure}
    \begin{subfigure}[t]{0.48\linewidth}
        \includegraphics[width=\textwidth]{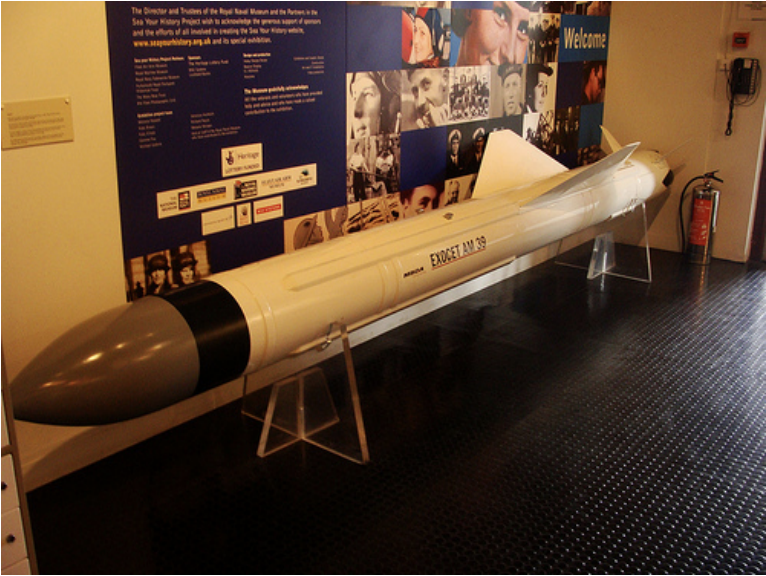}
        \caption{missile}
    \end{subfigure}%
    \caption{{\bf Examples of confusing labels in ImageNet \cite{tsipras2020imagenet}}. The labels above appear to correctly describe the visual concepts, however, ImageNet assigns \textit{acoustic\_guitar} and \textit{projectile} labels to the images, respectively.} 
    \label{fig:imagenet_confusion}
\end{figure}

We further elaborate on why it is important to align textual features on each downstream task. Previous work \cite{ren2022rethinking} has shown that CLIP's zero-shot transfer is vulnerable to expansion of downstream vocabulary used for class labels. This becomes even more important when the visual concepts in the target domain get associated with different class labels, presented at different granularities. 
\Cref{fig:imagenet_confusion} shows an example (taken from \cite{tsipras2020imagenet}) where multiple, different labels from ImageNet can be used to describe the same image.
Such cases are particularly difficult for vision-language models to generalize to in zero-shot manner, unless more context is given by either prompts or some domain-specific training data.

\begin{table}[b]
    \centering
    \begin{tabular}{l@{\hskip 0.1in}c@{\hskip 0.1in}c@{\hskip 0.1in}c@{\hskip 0.1in}c@{\hskip 0.1in}c@{\hskip 0.1in}c@{\hskip 0.1in}c}
    \toprule
    Few-shots &\ 1\ &\ 2\ &\ 4\ & \ 8\ &\ 16\\
    \midrule
    Linear-probe CLIP &22.17 &31.98 &41.20 &49.52 &56.13\\
	CoOp & 57.15 &57.81 & 59.99  &61.56 &62.95\\
	CLIP-Adapter & 61.20 & 61.52 & 61.84 &62.68 &63.59\\
	Tip-Adapter & 60.70 & 60.92 & 60.95 & 61.48 &62.00\\
    Tip-Adapter-F & 61.19 & 61.75 & 62.48 & 63.84 &65.47\\
    \midrule
    \textbf{DAC-V} & 60.71 & 61.48 & 61.87 & 63.38 & 64.89 \\
    \textbf{DAC-VT} & \textbf{61.32} & \textbf{62.39} & \textbf{63.11} & \textbf{64.78} & \textbf{66.61} \\
    \bottomrule
    \end{tabular}
    \caption{Top1 accuracy of different methods on ImageNet at different shots.}
    \label{tab:imgnet-comparisons-all}
\end{table}

Note that the adaptation of textual representations introduced in \cf \ref{sec:text_alignment} aims to caters for such confusing examples as it modulates the overall textual embedding (including the class name). Such an optimization allows the textual cache to adapt the class description according to the visual concepts defined by a few observed images.

\section{Understanding Inter-Modal and Intra-Modal Representations Alignment}
\begin{figure}[h]
    \centering
    \begin{subfigure}[t]{0.8\linewidth}
        \includegraphics[width=\textwidth]{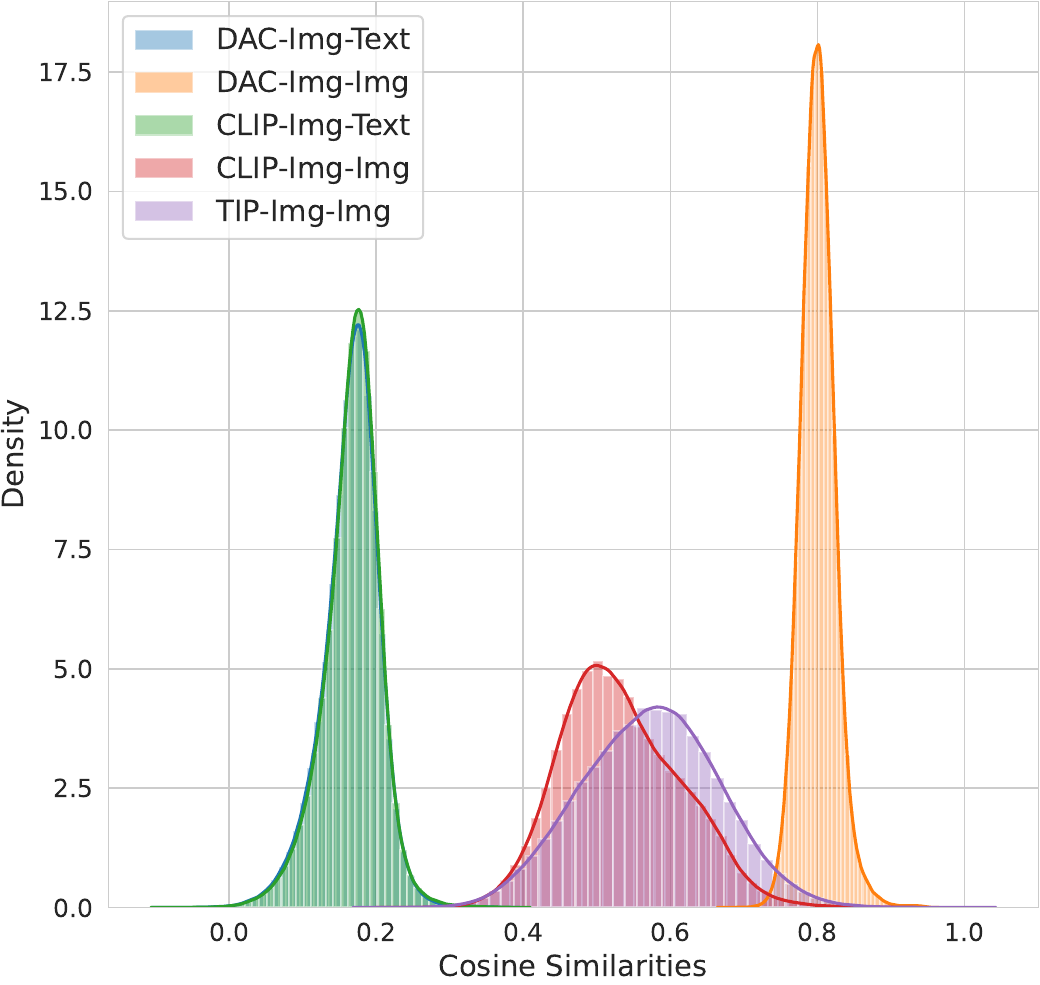}
    \end{subfigure}
    \caption{{\bf Pictorial depiction of modality gaps between intra-modal and inter-modal representations of different methods} (illustrated by cosine similarities). It can be seen that the DAC-VT's and CLIP image-text similarities remain within the same range.} 
    \label{fig:modality_gap}
\end{figure}

In this section, we delve into understanding how DAC-VT modulates the interactions between inter-modal and intra-modal representations. We look at them from the perspective of cone effects occurrences in representations distances that's been extensively studied in \cite{liang2022mind}. In \cref{fig:modality_gap}, we showcase the range of cosine similarities scores obtained by computing similarities between inter-modal and intra-modal representations. It can be seen that even after updating textual representations, DAC-VT maintains the same range of inter-modal similarity between images and text as in CLIP. The bigger shift is observed in intra-modal alignment where the visual representations tuned with DAC have a different support in comparison to TIP and CLIP based intra-modal alignments. We conjecture that this shift happens because the supervised contrastive objective used to tune visual representations introduce a different learning inductive bias than what was used to aligning image-text representations.

\end{document}